\RequirePackage{fix-cm}
\documentclass[twocolumn,natbib]{svjour3}          
\smartqed  
\usepackage{graphicx}
\usepackage{amsmath}
\usepackage{amsfonts}
\usepackage{booktabs}       

\usepackage{xcolor}

\usepackage{algorithm2e}
\RestyleAlgo{ruled}
\usepackage{multirow} 
\usepackage{float} 

\def\ie{{\em i.e.}}
\def\eg{{\em e.g.}}

\newcommand{\figref}[1]{Fig. \ref{#1}}
\newcommand{\tabref}[1]{Tab. \ref{#1}}

\newcommand{\secref}[1]{Section \ref{#1}}

\newcommand{\mc}[1]{\mathcal{#1}}

\newcommand{\br}[1]{\bm{\mathrm{#1}}}
\newcommand{\bs}[1]{\boldsymbol{\texttt{#1}}}

\usepackage{balance}

\usepackage[colorlinks, citecolor=blue]{hyperref}

\usepackage{bm}
\usepackage{bbm}
\usepackage{amssymb}

\journalname{}
\begin{document}
\title{Free Lunch to Meet the Gap: Intermediate Domain Reconstruction for Cross-Domain Few-Shot Learning
}


\author{Tong Zhang        \and Yifan Zhao$^{*}$ \and Liangyu Wang \and Jia Li$^{*}$
}


\institute{
$^{*}$: Corresponding author
\\
Tong Zhang \at
              State Key Laboratory of Virtual Reality Technology and Systems,
School of Computer Science and Engineering, Beihang
University, China. \\
              \email{tongzhang@buaa.edu.cn}           
           \and
Yifan Zhao \at
              State Key Laboratory of Virtual Reality Technology and Systems,
School of Computer Science and Engineering, Beihang
University, China. \\
              \email{zhaoyf@buaa.edu.cn}               
           \and
Liangyu Wang \at
              State Key Laboratory of Virtual Reality Technology and Systems,
School of Computer Science and Engineering, Beihang
University, China. \\
              \email{lyuwang@buaa.edu.cn}
           \and
Jia Li \at
              State Key Laboratory of Virtual Reality Technology and Systems,
School of Computer Science and Engineering, Beihang
University, China. \\
              \email{jiali@buaa.edu.cn} 
}


\maketitle

\begin{abstract}
Cross-Domain Few-Shot Learning (CDFSL) endeavors to transfer generalized knowledge from the source domain to target domains using only a minimal amount of training data, which faces a triplet of learning challenges in the meantime,~\ie, semantic disjoint, large domain discrepancy, and data scarcity. Different from predominant CDFSL works focused on generalized representations, we make novel attempts to construct \textbf{Intermediate Domain Proxies (IDP)} with source feature embeddings as the \textit{codebook} and reconstruct the target domain feature with this learned \textit{codebook}. We then conduct an empirical study to explore the intrinsic attributes from perspectives of \textit{visual styles} and \textit{semantic contents} in intermediate domain proxies. Reaping benefits from these attributes of intermediate domains, we develop a fast domain alignment method to use these proxies as learning guidance for target domain feature transformation. With the collaborative learning of intermediate domain reconstruction and target feature transformation, our proposed model is able to surpass the state-of-the-art models by a margin on 8 cross-domain few-shot learning benchmarks.


\end{abstract}

\section{Introduction}\label{sec1}

While current deep vision systems are undoubtedly successful at image classification tasks \citep{he2016deep, simonyan2014very}, their exceptional performance heavily relies on the availability of large-scale labeled data. Although these large-scale datasets \citep{deng2009imagenet} are making progress in facilitating networks to reach higher performance, it is usually impractical to gather such vast amounts of data when dealing with a novel concept. This data scarcity problem has motivated the research on Few-Shot Learning (FSL) \citep{fei2006one, lake2015human, miller2000learning}, which aims to model generalized memories from sufficient base training samples while tackling conceptually novel categories during the inference stage. 

Despite its substantial improvements in many ideal tasks, few-shot learning approaches suffer from a default assumption: \textit{the base pre-training categories for generalized knowledge and the few-shot novel categories should distribute in one same domain}. However, such strong assumptions are not feasible in most real-world systems, especially when exploring new concepts including remote-sensing satellite images~\citep{helber2019eurosat} and medical images~\citep{rajpurkar2017chexnet}. To meet the gap in realistic usage, Cross-Domain Few-Shot Learning (CDFSL) is established when various levels of domain distribution shifts exist between pre-trained base classes and target novel classes. Pioneer studies \citep{chen2019closer, guo2020broader} have demonstrated that predominant methods for FSL exhibit significant performance degradation when applied to the challenging problem of CDFSL. The huge gap between source and target domains impedes networks from extrapolating the generalized knowledge learned from source classes to target novel ones.

Recent approaches are devoted to learning a generalized representation to tackle this challenging problem, which generalizes the model from the source domain to the target domain. \cite{tseng2020cross} propose to augment the features with randomness by using the feature transformation layer. \cite{wang2021cross} augment multiple tasks in an adversarial manner. In comparison, \cite{liang2021boosting} introduce noisy distribution to enhance the network for learning robust image representations. Beyond these domain generalization methods, other research efforts~\citep{li2022ranking, shirekar2023self} focus on mitigating the domain gap from source to target ones with the facilitation of few-shot samples,~\eg, \cite{phoo2020self} propose to adapt the network to the target domain by performing self-supervised learning on the target domain. Nevertheless, this learning mechanism relies on the large additional amount of unlabeled data on the target domain, which is usually infeasible in certain practical scenarios.

One intuitive idea to address the CDFSL is to conduct few-shot learning with domain adaptation techniques,~\eg, MMD distance~\citep{tzeng2014deep, pan2010domain} or adversarial training~\citep{ganin2016domain, tzeng2017adversarial}. However, different from the prevalent domain issues, the CDFSL tackles a triplet of learning challenges. 1) \textbf{Semantic disjoint}: the semantic label spaces of the source and target domains are mutually exclusive, which is commonly shared in typical domain adaptation problems.  2) \textbf{Domain discrepancy} between domains can be extremely large, such as the stark contrast between the visual characteristics of natural images~\citep{deng2009imagenet} and X-ray images~\citep{wang2017chestx}. 3) \textbf{Data scarcity:} the $N$-way $K$-shot FSL samples are substantially difficult to represent the target domain distributions. These triplet simultaneous challenges in CDFSL lead to a clear failure when using prevalent domain alignment techniques.

Keeping these challenges in mind, in this paper, we make an attempt to construct Intermediate Domain Proxies, forming a shared latent space between the source and target domains. 
Toward this end, we first form a prototypical vector pool learned from the embedding of source categories and select the representative vectors to learn a mapping function from source to target features,~\ie, the dense source feature embedding serves as a codebook to reconstruct features of each target sample. We then transform the target samples with the learned mapping functions to construct an \textbf{intermediate domain}, which inherently follows two basic principles: 1) the intermediate domain shares the same \textit{semantic content} as the target domains; 2) the \textit{visual style} of intermediate domain achieves a good compromise between the source and target domains. Reaping benefits from these principles, the reconstructed features of the intermediate domain, namely proxies, exhibit fewer inherent domain gaps than their source counterparts while still retaining the visual cues of these sources. Instead of the naive alignment of source and target domains in prevailing works, we here use the intermediate domain proxies as the learning guidance to re-adjust both low and high-order parameters in feature normalization layers (\eg, Batch Normalization~\citep{ioffe2015batch}). Hence under extreme data scarcity, the network features can be fast aligned to the target domains. In conclusion, these intermediate proxies conduct relaxed alignment constraints instead of the direct alignment of target and source domains. During the domain alignment phase, we propose a rapid feature transformation using these proxies without the rehearsal of source data, and during the inference phase, our method does not rely on the constructed intermediate domains. This lightweight implementation indicates the "free-lunch" design of the proposed approach. The main contributions of our work are three-fold:

\begin{enumerate}
    \item We make attempts to construct Intermediate Domain Proxies (IDP) to solve the cross-domain few-shot learning problems and analyze the intrinsic attributes of these proxies from the perspective of visual styles and semantic content.
    \item We develop a fast adaptation method to use intermediate domain proxies as learning guidance for target domain feature transformations.
    \item We propose a unified framework for intermediate domain reconstruction and fast domain feature transformation in CDFSL. Experimental evidence indicates our proposed framework outperforms the state-of-the-art methods by a large margin on 8 public datasets.
\end{enumerate}

The remainder of this paper is organized as follows: ~\secref{sec:relatedwork} describes the related works of this research and ~\secref{sec:inter} presents an empirical study of the intermediate domain. ~\secref{sec:method} describes the proposed intermediate domain proxies reconstruction approach for the cross-domain few-shot learning problem. Qualitative and quantitative experimental results are reported in~\secref{sec:experiments}. ~\secref{sec:conclusion} finally concludes this paper.

\section{Related Works}\label{sec:relatedwork}

\textbf{Few-shot Learning} (FSL) aims to recognize novel concepts with very few numbers images, which are roughly categorized into two lines. The optimization-based approaches \citep{finn2017model, rusu2018meta, vuorio2019multimodal, li2017meta, nichol2018first} try to find a starting point for quickly optimizing the model. On the other hand, metric-based approaches  \citep{oreshkin2018tadam, snell2017prototypical, sung2018learning, vinyals2016matching, xu2022alleviating, chen2022iterative} tend to find a task-independent embedding space that can be generalized to the target category by designing metric functions. To better capture detailed features of the image, recent metric-based approaches \citep{zhang2020deepemd, wertheimer2021few, ye2020few} have focused on dense measuring of the feature map, rather than the global representation prototypes. Several methods \citep{rizve2021exploring, gidaris2019boosting, wei2022embarrassingly, luo2021rectifying} set auxiliary tasks in the pre-training phase to enhance the model's generalization ability. \cite{doersch2020crosstransformers} utilize the attention scheme to transfer the pretrained knowledge to few-shot learning. Nevertheless, these methods do not consider the significant differences between the source and target domains in the CDFSL setting.

\textbf{Domain Adaption} methods \citep{tzeng2017adversarial, cui2020gradually, robey2021model, kang2018deep, zhang2019bridging} align the source and target domains to resolve domain shifts. These methods usually focus on regularizing the feature similarity of source and target domains, thus confusing the backbone networks to construct a unified representation for both domains.
Another line of methods \citep{gong2012geodesic, gopalan2013unsupervised, dai2021idm} propose reducing the domain alignment difficulty by designing intermediate domains to connect the source and target domains. 
Unlike these methods, the CDFSL task in this paper requires solving massive target domain classification tasks simultaneously with access to only a tiny number of samples from the supporting dataset. Simply mitigating the domain gaps would lead to severe overfitting on these few-shot samples.

\textbf{Cross-Domain Few-shot Learning} (CDFSL) methods \citep{guo2020broader, wang2021cross, phoo2020self, liang2021boosting} usually adopt the framework of transfer learning,~\ie, supervised learning using source domain data and then fine-tuning using target domain samples. This simple pipeline has proven superior to many SOTA FSL methods in CDFSL settings \citep{chen2019closer, guo2020broader}.  Apart from these works,~\cite{li2022cross} focus on cross-domain domain generalization and fast adaptation to unseen domains with network tuning techniques, which stands at a different view from the conventional CDFSL learnings. Besides, \cite{xu2021memrein} propose a contrastive learning scheme to distill the memorized knowledge from source domains. 
Representative methods~\citep{phoo2020self, li2022ranking, shirekar2023self} suggest enhancing cross-domain discrimination through either pseudo-labeling for model distillation or GNN-based message passing in the target domain. 
However, in these works~\citep{phoo2020self, li2022ranking, shirekar2023self}, unlabeled target domain data are not always available in practical tasks,~\eg, X-ray images. Unlike these methods, we argue that in the fine-tuning phase, the network only has access to the target domain data is not the most effective use of the knowledge learned in the past due to catastrophic forgetting \citep{mccloskey1989catastrophic, kemker2018measuring}. In contrast, in this paper, we propose to find a solution that only uses few-shot target domain data and without the rehearsal from the large-scale source domain data. With this in mind, we propose to build intermediate domain proxies instead of accessing additional source or target data.

\textbf{Discussions and Relations.} The concept of intermediate domain is proposed in domain adaptation and extended to many applications in previous studies, including \cite{gong2012geodesic, gopalan2013unsupervised, dai2021idm}. However, these works require a huge demand for training samples of the target domain, which is unavailable for the few-shot scenarios.  Thus we resort to the closed-form feature reconstruction methods~\citep{bertinetto2018meta,wertheimer2021few} to build intermediate domains in the feature space using dense prototypical sources~\citep{snell2017prototypical}. By leveraging the advantage of the intermediate domain and feature reconstruction, the intermediate proxies are generated as a bridge when there are extremely few-shot available samples in target domains. Based on these proxies, beyond the conventional feature space alignment, our approach conducts a fast domain adaptation by normalization feature transformation techniques.

\begin{figure*}[!tbp]
    \centering
    \includegraphics[width = \textwidth]{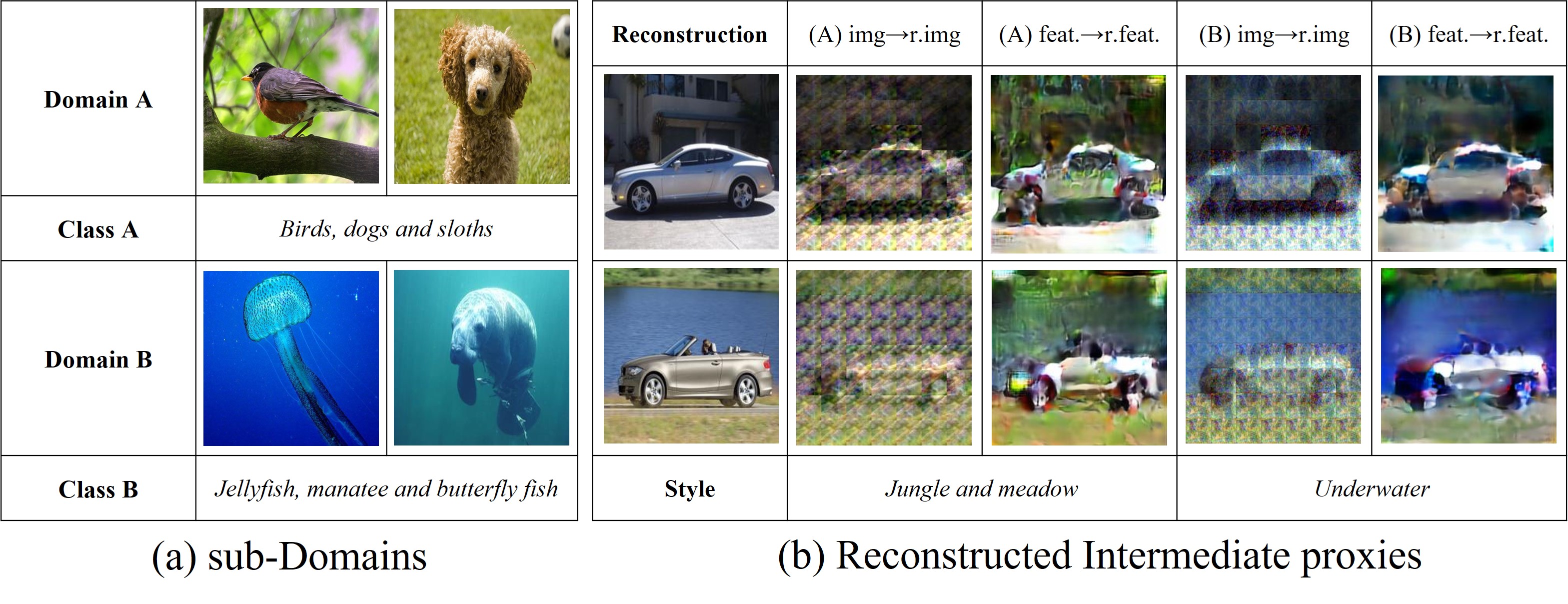}
    \caption{Illustration of intermediate proxy reconstruction. a) Two representative sub-domains for reconstruction bases. b) Reconstructed intermediate proxies using the sources in a).}
    \label{fig:visualization_ia}
\end{figure*}

\section{Intermediate Domain Reconstruction: An Empirical Study}\label{sec:inter}
\subsection{Motivations and Setup}
Imagine we are drawing an object, we almost always first draw the \textit{sketch} of the object and then fill in the color with various \textit{pigments and inks}, finishing with wash, canvas, or watercolor styles. Analogous to this painting process, in CDFSL, we decouple the object representation into \textbf{semantic contents} and \textbf{visual styles}. Our motivation is to find an intermediate proxy domain $\mc{P}$ that shares the same semantic features of samples in the target domain $\mc{T}$ by using the base units in the source domain $\mc{S}$, forming the intermediate domain with a remix of both domains. 


\textbf{Empirical Study.} Toward this motivation, we start from constructing the intermediate proxies with few visually similar images sampled from two ``sub-Domains" in the~\textit{mini}ImageNet ~\citep{vinyals2016matching},~\ie, i) Domain $\mc{A}$: \textit{objects in jungle and meadow} including birds, dogs, and sloths; ii) Domain $\mc{B}$: \textit{underwater objects} such as Jellyfish. We then take one exemplar image from Car dataset~\citep{krause20133d} as the target ``domain", which shows different contents that never appeared in the source domain, shown in~\figref{fig:visualization_ia}. 

\textbf{Construction of Intermediate Proxy} Suppose 
one domain as source domain $\mc{S}$ and the other as target domain $\mc{T}$, we first extract feature bases $\{\br{C}_{i}\}_{i=1}^{n} \in \mathbb{R}^{n\times d} $ by network backbone $f_{\theta}$, where $n$ denotes the number of features and $d$ denotes their dimensions. To construct intermediate proxies, one intuitive method is to reconstruct the feature maps of each target sample using the elements in the source bases, serving as the codebook. Hence the reconstructed feature map could share both the source and target domain features,~\ie, reconstructed bases are similar to source domain $\mc{S}$ and reconstructed goals are similar to target domain $\mc{T}$. In this paper, we name this process as the Intermediate Proxy Construction.

Let $\br{T}\in \mathbb{R}^{r\times d}$ denote the target domain embeddings required to be reconstructed. Following the Ridge Regression~\citep{wertheimer2021few,hoerl1970ridge, bertinetto2018meta}, and Sparse Coding mechanisms~\citep{mairal2010online}, hence we need to find a mapping matrix $\br{W}\in \mathbb{R}^{r\times n}$ to minimize the reconstruction error $\sum_{j=1}^{r} ||\br{T}_{j}- \br{W}\br{C} ||^{2}_{2}$. To retain this reconstruction as a convex process, we have the following Ridge regression form:
\begin{equation}
  \widehat{\br{W}} = \arg{\min\limits_{\br{W}}\left\| {\br{T} - \br{W}\br{C}} \right\|^2_2} + \lambda\left\| \br{W} \right\|^2_2.
  \label{eq:reconstruction}
\end{equation}
where $\lambda$ is the hyper-parameter to balance the regularization of $\ell_2$-norm. With this regularization, we hereby use its closed-form solution as in~\citep{bertinetto2018meta,wertheimer2021few}, which also prevents the low-rank issues ($n<r$) in solving Eq. (1), ~\ie, making $\widehat{\br{W}}$ an invertible matrix:
\begin{equation}
  \br{P} = \widehat{\br{W}}\br{C} = \br{T}\br{C}^{\top}(\br{C}\br{C}^{\top}+\lambda \br{I})^{-1}\br{C}.
  \label{eq:solution}
\end{equation}
Here reconstructed intermediate proxies $\br{P} \in \mathbb{R}^{r\times d}$ have the same size as target domain features $\br{T}$.

\begin{figure*}[!t]%
\centering
\includegraphics[width=0.7\textwidth]{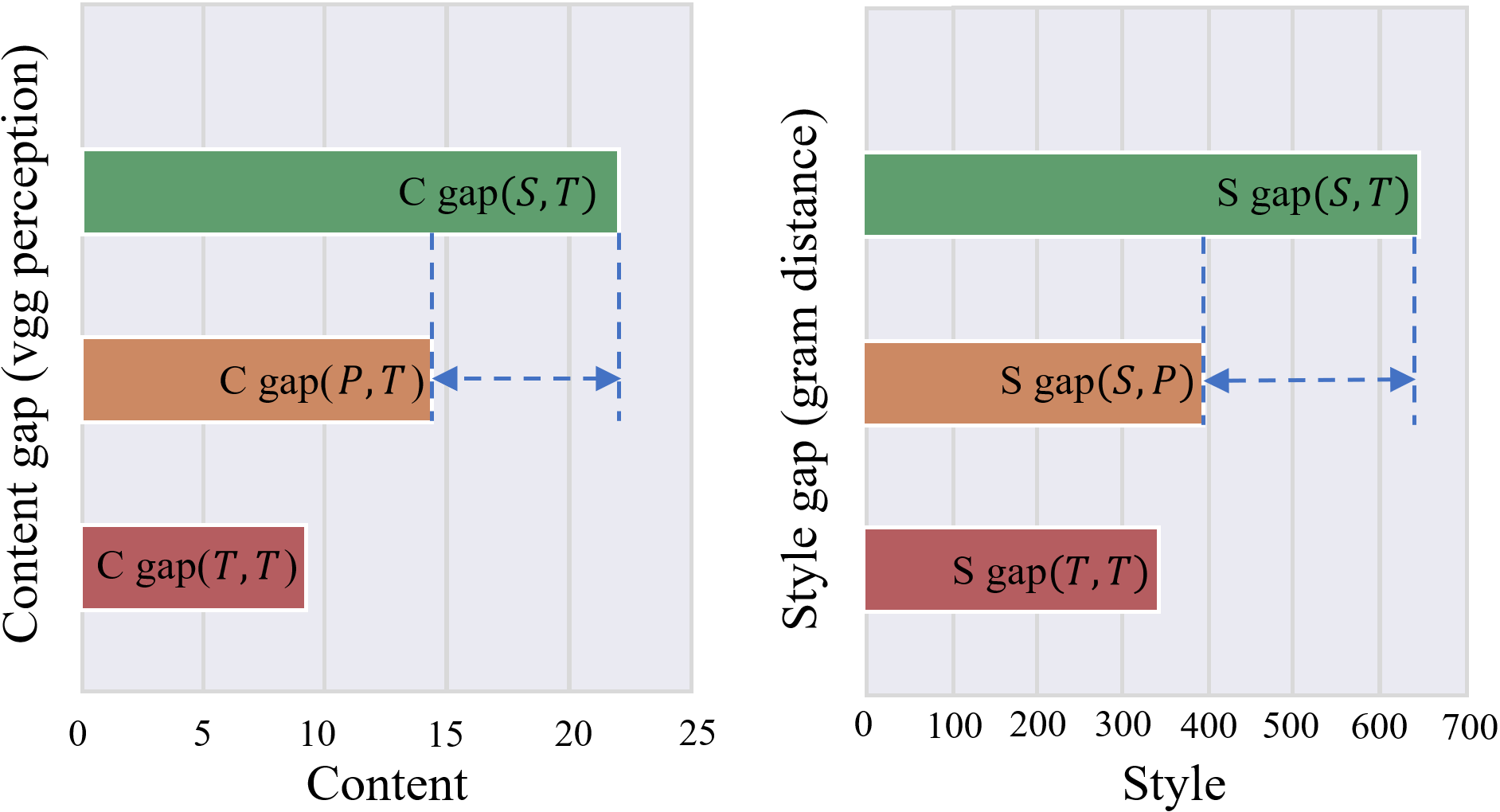}
\caption{Quantitative evaluations of content and style differences among source domains $\mc{S}$, target domains $\mc{T}$, and intermediate domain proxies $\mc{P}$. The content distance is calculated by VGG perception and Style distance is calculated by Gram distance using samples in \textit{
mini}-ImageNet (Detailed in Appendix A). 
} 
\label{fig:visualization_ib}
\end{figure*}

\subsection{Analyzing Styles and Semantics of Intermediate Proxy}
With the reconstructed intermediate proxies, here arise two inherent questions. \textbf{Q1}: \textit{How does the choice of base $\{\br{C}_{i}\}_{i=1}^{n}$ impact the domain reconstruction?} and \textbf{Q2}: \textit{What is the relationship of intermediate proxies with the source/target domains?} 

To answer these two questions, here we conduct two lines of reconstruction with quantitative and qualitative analyses. $\bs{img} \xrightarrow{} \bs{r.img}$ denotes that we use source images to learn mapping weights $\widehat{\br{W}}$ on target images. And $\bs{feat} \xrightarrow{} \bs{r.feat}$ denotes that we use the feature reconstruction process as in Eqs.~\eqref{eq:reconstruction} and \eqref{eq:solution}. The feature extraction process with $f_{\theta}$ uses the pretrained ResNet-10 backbone and other implementation details could refer to~\secref{sec:imp} and Appendix A. In~\figref{fig:visualization_ib}, we calculate the content distance (VGG perceptional scores~\citep{simonyan2014very}) and style variances (Gram distance~\citep{gatys2015neural} of among source $\mc{S}$, target $\mc{T}$ and intermediate proxies $\mc{P}$.  Here we draw three principles:

\begin{enumerate}
\item \textbf{The Intermediate Proxy preserves the semantic content of the target image.}  In~\figref{fig:visualization_ia}, the reconstructed images clearly shows the similar content~\ie, the cars with the target domain images. While the distance of content of $(\mc{P},\mc{T})$ is much more closer than $(\mc{S},\mc{T})$.

\item \textbf{The Intermediate Proxy reflects the source domain style when using different bases.} Comparing the domain $\mc{A}$ and $\mc{B}$, the reconstructed images show {\color[RGB]{96,177,87} \textbf{green}} background when using jungle images while exhibiting {\color{blue}\textbf{blue}} backgrounds using underwater images as bases.  

\item \textbf{The Intermediate Proxy shows fewer domain shifts in content and style statistics than source domains.} Both the style and content shift in $(\mc{P},\mc{T})$ are much less than $(\mc{S},\mc{T})$ in~\figref{fig:visualization_ib}, indicating aligning with intermediate domain proxies $\mc{P}\rightarrow \mc{T}$ would be much easier than the direct alignment of source and target ones $\mc{S}\rightarrow \mc{T}$. Note that the distance to target domains could be closer if more base vectors are used for reconstruction.
\end{enumerate}

\subsection{The Role of Intermediate Proxy}
Besides the intuitive visualization and experimental statistics, here we provide detailed theoretical analyses of why the intermediate proxies help cross-domain learning.

\textbf{Definition 1}\textit{ (Inter-domain discrepancy distance).} 
\textit{Inter-domain discrepancy distance between the source domain $\mathcal{S}$ and target domain $\mathcal{T}$ is measured by the Euclidean distance of their corresponding sample features: $disc_\mathcal{L}(\mathcal{S}, \mathcal{T}) = \sum_{ij} \left\| \mathcal{S}_i - \mathcal{T}_j \right\|^2$, with smaller discrepancy distances indicating greater semantic similarity between the domains.}

\textbf{Proposition 1}\textit{ (High semantic similarity).}
\textit{By controlling the ridge regression regular term $\lambda$ , the semantic similarity between the intermediary domain proxy $\mathcal{P_\lambda}$ and the target domain $\mathcal{T}$ is larger than that between the source domain $\mathcal{S}$ and the target domain $\mathcal{T}$. Their inter-domain discrepancy distance satisfies the following relationship:}
$\exists~\lambda$ \textit{,s.t.} $disc_\mathcal{L}(\mathcal{S}, \mathcal{T}) > disc_\mathcal{L}(\mathcal{P}, \mathcal{T})$.

The first proposition indicates that the intermediate domain $\mathcal{P}$ shares more semantic similarity with the target domain $\mathcal{T}$, thus aligning the intermediate domain is much easier than the direct alignment of source and target domains. This proposition provides a compromised solution when facing an extremely huge domain gap or when there is rare data for aligning these domains.   

\begin{figure*}[tb]
    \centering
    \includegraphics[width = 1.0\textwidth]{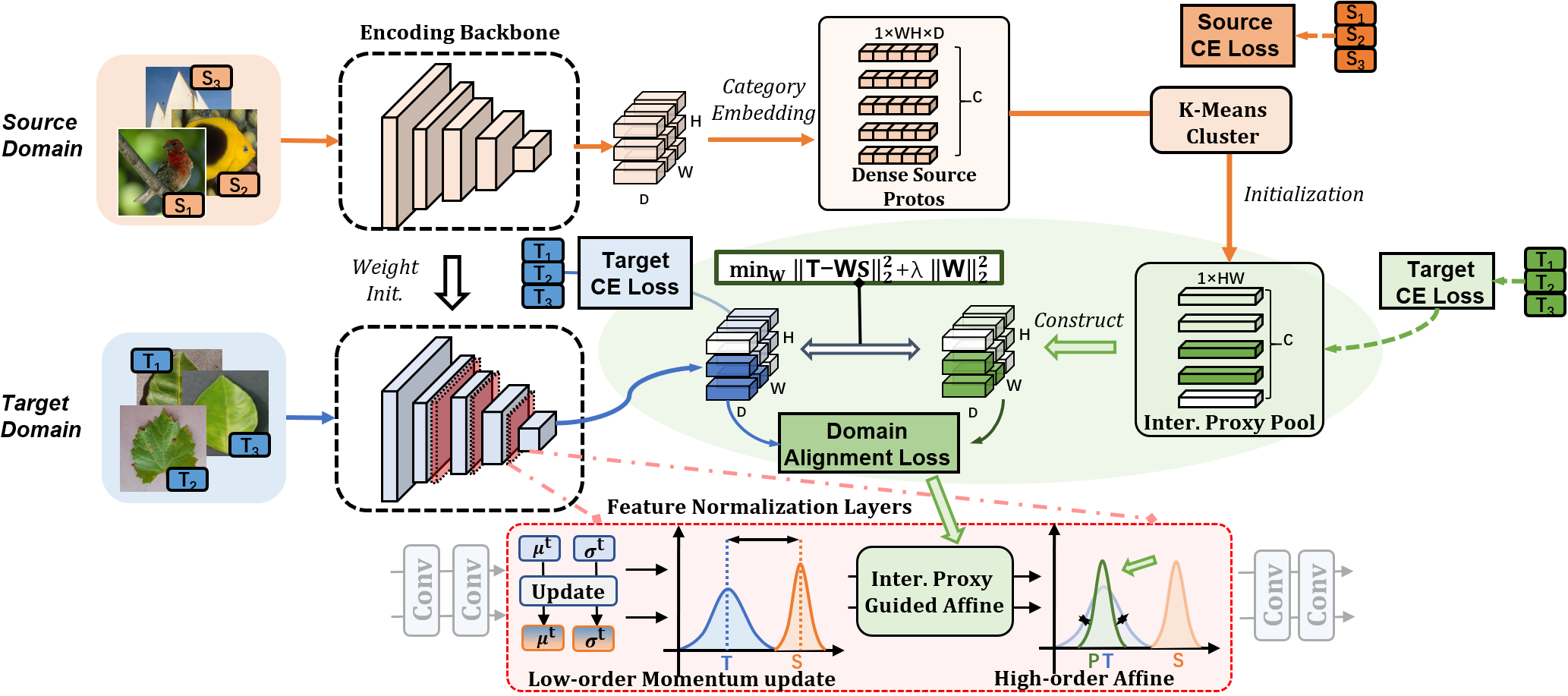}
    \caption{Illustration of the proposed method. We first collect dense prototypes from source domains and use them to construct the intermediate proxy pool. Features in this pool are then employed to reconstruct the target domain, forming intermediate reconstructions. After that, the intermediate proxies are adopted as learning guidance for fast feature alignment of target and intermediate domains.}
    \label{fig:pipeline}
\end{figure*}

\textbf{Proposition 2} \textit{(Reducing target classification error).}
\textit{Aligning the target domain $\mathcal{T}$ to the intermediate domain proxy $\mathcal{P}_\lambda$ can reduce the discrepancy distance between the source and target domain $disc_\mathcal{L}(\mathcal{S}, \mathcal{T})$, which in turn reduces the error of the classifier $\epsilon_\mathcal{T}$ on the target domain.}

Besides the first proposition, the other crucial issue is that aligning the intermediate domain and target domain should in turn reduce the gap between the source and target domains. Thus we could regard the intermediate domain as a substitution during the optimization.
Please refer to the Appendix for detailed proofs.

\section{Approach}\label{sec:method}

\subsection{Dense Reconstruction for Cross-domain Few-shot Learning}
\textbf{Problem Formulation.} Given the source and target domain  $\mathcal{D}_s$ and $\mathcal{D}_t$,  the distribution $\mc{N}(\cdot)$ of $\mathcal{D}_s$ and $\mathcal{D}_t$ are in different semantic space,~\ie, $\mc{N}(\mathcal{D}_s) \nsim \mc{N}(\mathcal{D}_t)$. The large-scale labeled images of base classes $C_{base}$ are available in the source domain,~\ie, $\mathcal{D}_s=\{(\mathbf{x}_i^s,\mathbf{y}_i^s)\}_{i=1}^L$, where $L$ is the number of images in $\mathcal{D}_s$. For $N$-way $K$-shot problems, the target domain $\mathcal{T}$ includes two parts: a support dataset $\mathcal{T}^{s}=\{(\mathbf{x}_i^{ts},\mathbf{y}_i^{ts})\}_{i=1}^{K\times N}$ with a few labeled samples and a unknown query dataset $\mathcal{T}^{q}=\{(\mathbf{x}_i^{tq},\mathbf{y}_i^{tq})\}_{i=1}^M$ for inference. Here $N, K$ denotes the number of classes and images in each class in $\mathcal{T}^{s}$. The label space of source domain $\mathcal{D}_s$ is disjoint from target domain $\mathcal{T}$, ~\ie, $C_{base} \bigcap C_{novel} = \phi$. The optimization objective follows the conventional FSL problems, which aim to learn a generalized embedding from base classes and then transfer it to target domains with few-shot data. The optimized parameters are typically composed of two major components: 1) one backbone network $f_{\theta}$ to encode the images into dense feature maps; 2) one classifier $g_{\mc{V}}$ to predict probabilities for each category. Thus we have 
\begin{equation}\label{eq:formulation}
\begin{aligned}
    \mathcal{L}_{ce}(\theta,\mc{V}) &= 
    \mathbb{E}_{(\mathbf{x}_i^o, \mathbf{y}_i^o) \sim \mathcal{D}_{o}} 
    [\mathbf{y}_i\log(g_\mc{V}(\bs{Pool}(f_\theta(\mathbf{x}_i^o))))], \\& o \in \{s,t\}.
\end{aligned}
\end{equation}

\textbf{Dense Reconstructions for CDFSL.} The conventional learning schemes in Eq.~\eqref{eq:formulation} construct classifiers following a prototypical trend~\citep{snell2017prototypical},~\ie, $\br{V} =\sum_{i=1}^{k} \sum_{j=1}^{WH}(f_{\theta}(\br{x}_{i,j})) $ for given $k$ samples and spatial dimension $j$. $W,H$ are the width and height of the feature map. Although this learning scheme provides satisfactory feature embeddings in common FSL problems, it still suffers two major challenges in cross-domain learning: 1) the spatial information is severely neglected by pooling operations, while for novel categories in the target domain, representing novel objects with only global prototypes are usually difficult; 2) global prototypes provides more domain-specific semantics while losing generalization when serving as materials for reconstruction. Following the reconstruction process in Eq.~\eqref{eq:reconstruction} and prevailing works~\citep{bertinetto2018meta,wertheimer2021few}, we here first reconstruct the source domains with prototypes $\br{V}^{s}$ without losing its spatial dimension. This reconstruction measurements $\mathcal{R}_{\mc{A}\rightarrow \mc{B}}(\cdot)$ by using domain $\mc{A}$ to construct $\mc{B}$ also follows a closed form solution with learnable $\br{V}^{a}$:
\begin{equation}\label{eq:sourcerec}
    \mathcal{R}_{\mc{A}\rightarrow \mc{B}}(f_\theta(\mathbf{x}^{b});\br{V}^{a})=\frac{\exp({||\br{W}^{a}\br{V}^{a}_i-f_\theta(\mathbf{x}^{b})}||^2)}{\sum_{j=0}^{N-1}\exp{(||\br{W}^{a}\br{V}^{a}_j-f_\theta(\mathbf{x}^{b})||^2})},
\end{equation}
where $\br{W}^{a}=f_\theta(\mathbf{x})\br{V}^{a^\top}(\br{V}^{a}\br{V}^{a^\top}+\lambda\br{I})^{-1}$ denotes the inter-domain mapping weights. $a,b$ denotes the matrices or vectors belonging to domains $\mc{A}$ and $\mc{B}$ respectively. We thus form a reconstruction $\mc{R}_{\mc{S}\rightarrow \mc{S}}(\cdot )$ by using source domain bases to construct itself as pretraining to get generalized embedding. After this pretraining stage in only source domains, the densely formed visual prototypes $\br{V}$ can be constructed and the network parameters are initialized for subsequent cross-domain learning. Note that the reconstruction process for prototypes mainly follows~\cite{wertheimer2021few} to optimize the few-shot process in a differentiable manner. 

\subsection{Intermediate Domain Construction}
\label{sec:domain_alignment}

\textbf{Intermediate Proxies Generation.} Starting from Eq.~\eqref{eq:sourcerec}, we then have a densely formed visual prototypes $\br{V} \in \mathbb{R}^{WH \times D\times C_{base}}$, where $C_{base},D$ denotes the number of base classes and feature dimensions. The major concern is what we asked in \textbf{Q1} in~\secref{sec:inter},~\ie, how to choose reconstructed materials for the intermediate domain without additional costs? Using base pooling prototypes $\mathbb{R}^{WH \times D\times C_{base}}$ with such high dimension for reconstruction would lead to huge computation costs and inferior optimization for the closed-form equation in Eq.~\eqref{eq:reconstruction}. Hence we use the K-means algorithm to cluster the dense prototypes into spatial feature pool $\mc{U}$ for intermediate domain construction $\mc{P}$, resulting in a cluster mapping $\mc{M}: \mathbb{R}^{WH \times D\times C_{base}}\rightarrow \mathbb{R}^{WH \times D}$, as in~\figref{fig:pipeline}. This clustering process also regularizes different semantic classes with the unified reconstruction materials. Different from prevailing FSL efforts, the intermediate proxy pool $\mc{U}=\{\br{U}_i \}_{i=1}^{WH}$ depicts a generalized representation of each visual pattern, thus is inherent to reconstruct further novel categories with large domain gaps.` By replacing the generalized learnt $\br{U}$ with $\br{C}$ that only for specific samples, for each category $n\in [1,N]$, we have 
\begin{equation}
  \br{P}^{i} = \widehat{\br{W}}^{i}\br{U} = \br{T}^{i}\br{U}^{\top}(\br{U}\br{U}^{\top}+\lambda \br{I})^{-1}\br{U} \in \mathbb{R}^{K WH\times D},
  \label{eq:gen}
\end{equation}
where $\br{T}^{i}$ indicates the $i$th class of the target domain features. Eq.~\eqref{eq:gen} means different classes in the novel set share similar reconstruction materials from the proxy pool $\mc{U}$. We perform reconstruction for each category independently to obtain its respective intermediate domain. This observation leads us to an interesting manner to further tune these proxies and make them fully adapted to target domains,~\ie, the constructed proxy $\br{P}^{i}$ highly corresponds with the supported features $\br{T}^{i}$ for class ${C}_{i}$. Hence we use the standard cross entropy to retain the reconstructed proxies in the target domain semantic space with $C_{novel}= N$ categories:
\begin{equation}\label{eq:poolconstraint}
    \mathcal{L}_{proxy}(\theta)= 
    \mathbb{E}_{(\mathbf{x}_i^{ts}, \mathbf{y}_i^{ts}) \sim \mathcal{D}_t} 
    [\mathbf{y}^{ts}_i\log(\mathcal{R}_{\mc{P}\rightarrow \mc{T}}(f_{\theta}(\br{x}),\br{P}_{i}))],
\end{equation}
where $\mathcal{R}_{\mc{P}\rightarrow \mc{T}}$ denotes the reconstruction function from proxies $\mc{P}$ to target domains $\mc{T}$. Note that during this process, the proxies are not learnable matrices and can only be updated indirectly by the change of extracted source prototypes $\check{\br{V}}^{s}$, and then be re-mapped by  $\{\check{\br{U}}\}_{i} =\mc{M}(\check{\br{V}}^{s}) $.

\subsection{Domain Alignment with Intermediate Proxies}

After constructing intermediate proxies distributed between source and target domains, one intuitive idea is to align target and intermediate domains $(\mc{P},\mc{T})$ instead of the direct alignment of $(\mc{S},\mc{T})$. Beyond this consideration, we observed that due to the extremely small size of training samples, this compromised alignment by using intermediate domains still leads to catastrophic overfitting on very few target samples. To overcome this, here we propose to transform the feature statistics beyond tuning the whole network. 

\textbf{Decomposing Batch Normalization.} Conventional BN layers~\citep{ioffe2015batch} are typically decomposed into two stages, statistic normalization and affine transformations.  Given an extracted feature map $\br{F} \in \mathbb{R}^{B \times H \times W \times D}$ with the batch-wise dimension, the mean and variances are calculated along the channel size 
\begin{equation}
\mu_{\bs{BN}} = \frac{1}{B \times H \times W} \sum_{b=1}^{B} \sum_{r=1}^{H \times W} \br{F}_{b,c,r},
\end{equation}
and the channel-wise variance is computed as 
\begin{equation}
\sigma^2_{\bs{BN}} = \frac{1}{B \times H \times W} \sum_{b=1}^{B} \sum_{r=1}^{H \times W} (\br{F}_{b,c,r} - \mu_{\bs{BN}})^2. 
\end{equation}
Thus the feature normalization in Eq.~\eqref{eq:f_bn} indicates the \textbf{low-order statistics}~\citep{maria2017autodial, wang2019transferable}, which are mainly decided by the historical statistics and current running status. While Eq.~\eqref{eq:f_aff} forms a \textbf{high-order} affine transformation to adjust the distribution shape with the scaling factor $\gamma$ and shifting factor $\beta$.
\begin{equation}
\label{eq:f_bn}
\br{F}_{\bs{BN}}=\frac{\br{F}-\mu_{\bs{BN}}}{\sqrt{\sigma_{\bs{BN}}^2+\epsilon}},
\end{equation}
\begin{equation}
\label{eq:f_aff}
\br{F}_{\bs{aff}}=\gamma \br{F}_{\bs{BN}}+\beta.
\end{equation}

\textbf{Feature Transformation with Intermediate Domain Proxies.}
Dozens of research efforts~\citep{li2016revisiting} have demonstrated the strong correlation between visual styles and feature statistics and several pioneer works~\citep{zhang2020generalizable} focus on the BN optimization in domain adaptation problems. Inspired by these early explorations, here we resort to the intermediate proxies for the fast alignment of multiple domains. First, the low-order statistics in the normalization layers are updated by a static momentum,~\ie, updating the same ratio of samples during different training phases. Here we argue to construct a dynamic momentum function that is gradually increased during the training phase. 
\begin{equation}
\Tilde{\mu}_t=(1-\mc{G}^\alpha(t))\cdot \Tilde{\mu}_{t-1} + \mc{G}^\alpha(t) \cdot \mu_t,
\label{eq:mu}
\end{equation}
\begin{equation}
\Tilde{\sigma}_t^2=(1-\mc{G}^\alpha(t))\cdot \Tilde{\sigma}_{t-1}^2 + \mc{G}^\alpha(t) \cdot \sigma_t^2,
\label{eq:sigma}
\end{equation}
where $\mc{G}^\alpha(t)=1/(1 + \exp(-t/\alpha))$, $\alpha$ is used to control the scale of weighting function. $\Tilde{\mu}_t,\Tilde{\sigma}_t$ denotes the updated mean and variances for time step $t$.
Thus in the first training stages, the models tend to use more source statistics for representation stability.

\begin{algorithm}[!t]
    \LinesNumbered
    \caption{Cross-domain Few-shot Learning with Intermediate Domain Proxies (IDP)}
    \label{algo:IDP}
    \KwIn{Source domain data $\mc{S}$, target domain support set $\mc{T}^s$, target domain query set $\mc{T}^q$.}
    \KwOut{Backbone network $f_\theta$, Classifier $g_\mc{V}$, Predicted results $\br{S}$.}
    Init.  parameters of the backbone network $\theta$ in $F_\theta(\cdot)$ with random norm\;
    \tcp{Source Domain Pretraining}
    Random Init. source domain densely formed  visual prototypes: $\br{V}^s\in \mathbb{R}^{WH\times D\times C_{base}}$\;
    \For{$\forall(\br{x}^s,\br{y}^s)\in \mc{S}$}{
        Calculate inter-domain mapping weights $\br{W}^{s}=f_\theta(\mathbf{x})\br{V}^{s^\top}(\br{V}^{s}\br{V}^{s^\top}+\lambda\br{I})^{-1}$\;
        Calculate reconstruction measurements $\mathcal{R}_{\mc{S}\rightarrow \mc{S}}(f_\theta(\mathbf{x});\br{V}^{s})$ by Eq. \ref{eq:sourcerec}\;
        Calculate cross-entropy loss $\mathcal{L}_{ce}$ by Eq. \ref{eq:formulation} and optimizing $\theta, \mc{V}=\arg\min_{\theta, \mc{V}}\mathcal{L}_{ce}$ \;
    }
    \tcp{Target Domain Finetuning}
    Init. parameters of the backbone network $\theta$ in $F_\theta(\cdot)$ with pre-training\;
    \For{fine-tining time step $t$ and $\forall(\br{x}^{ts},\br{y}^{ts})\in \mc{T}^s$}{
        Calculate cross-entropy loss for target domain finetuning $\mc{L}_{tar}$\;
        Cluster the dense prototypes $\br{V}^s$ into spatial feature pool $\mc{U}$\;
        Reconstruct intermediate proxies $\br{P}^i$ using generalized learnt $\br{U}$ by Eq. \ref{eq:gen}\;
        Calculate cross-entropy loss for reconstructed proxies $\mc{L}_{proxy}$ by Eq. \ref{eq:poolconstraint}\;
        Update low-order statistics $\Tilde{\mu}_t$, $\Tilde{\sigma}_t^2$ by Eq. \ref{eq:mu} and Eq. \ref{eq:sigma}\;
        Calculate reconstruction measurements $\mc{R}_{\mc{P}\rightarrow \mc{T}}(\br{P};\br{V}^{t})$ by Eq. \ref{eq:sourcerec}\;
        Calculate intermediate proxies constraint loss $\mc{L}_{align}$ by Eq. \ref{eq:align}\;
        Gather total loss $\mc{L}_{sum}$ by Eq. \ref{eq:sum} and optimizing $\theta, \gamma,\beta=\arg\min_{\theta, \gamma,\beta}\mc{L}_{sum}$ \;
    }
    \tcp{Target Domain Querying}
    \For{$\forall(\br{x}^{tq},\br{y}^{tq})\in \mc{T}^q$}{
        Calculate reconstruction measurements $\mc{R}_{\mc{T}\rightarrow \mc{T}}(f_\theta(\br{x}^{tq});\br{V}^{t})$ by Eq. \ref{eq:sourcerec}\;
        Predicting the probability of each category $\mathbf{S}$ using $\mc{R}_{\mc{T}\rightarrow \mc{T}}$\;
    }
    \Return Predicted results $\br{S}$ of $N$ classes
\end{algorithm}

\begin{table*}[t]
  \centering
  \caption{Comparisons with state-of-the-art models on CDFSL benchmark dataset. The first and second best values on each dataset are highlighted in bold and underlined.
  $^{*}$: Finetuning using the same optimization as Ours. $^{\dag}$: re-trained using ResNet-12 as backbone, others using ResNet-10.  }
  \resizebox{\textwidth}{!}{
    \setlength{\tabcolsep}{0.3mm}
\renewcommand{\arraystretch}{1.3}
    \begin{tabular}{c|cccc|cccc}
    \toprule
    1-shot & ISIC  & EuroSAT & CropDisease & ChestX & Car   & CUB   & Plantae & Places \\
    \midrule
    GNN~\citep{garcia2018few}   & 32.02±0.66 & 63.69±1.03 & 64.48±1.08 & 22.00±0.46 & 31.79±0.51 & 45.69±0.68 & 35.60±0.56 & 53.10±0.80 \\
    FWT~\citep{tseng2020cross}   & 31.58±0.67 & 62.36±1.05 & 66.36±1.04 & 22.04±0.44 & 31.61±0.53 & 47.47±0.75 & 35.95±0.58 & 55.77±0.79 \\
    LRP~\citep{sun2021explanation}   & 30.94±0.30 & 54.99±0.50 & 59.23±0.50 & 22.11±0.20 & 32.78±0.39 & 48.29±0.51 & 37.49±0.43 & 54.83±0.56 \\
    AFA~\citep{hu2022adversarial}   & 33.21±0.30 & 63.12±0.50 & 67.61±0.50 & 22.92±0.20 & 34.25±0.40 & 46.86±0.50 & 36.76±0.40 & 54.04±0.60 \\
    STARTUP~\citep{phoo2020self} & 32.66±0.60 & 63.88±0.84 & 75.93±0.80 & 23.09±0.43 & -     & -     & -     & - \\
    TPN-ATA~\citep{wang2021cross} & 33.21±0.40 & 61.35±0.50 & 67.47±0.50 & 22.10±0.20 & 33.61±0.40 & 45.00±0.50 & 34.42±0.40 & 53.57±0.50 \\
    FRN$^{\dag}$~\citep{wertheimer2021few} & 33.73±0.62 & 63.80±0.91 & 71.93±0.85 & 22.52±0.40 & 32.37±0.58 &\textbf{51.76±0.80} & \underline{42.37±0.73}     & \textbf{56.92±0.84} \\
    FRN$^{*}$$^{\dag}$~\citep{wertheimer2021few}  & 33.38±0.58 & 60.25±0.81 & 70.09±0.82 & 22.53±0.38 & 33.08±0.57    & 44.95±0.74    & 36.45±0.65    & 50.84±0.75 \\
    ConFT~\citep{das2021importance} & \underline{34.47±0.60} & 64.79±0.80 & 69.71±0.90 & \textbf{23.31±0.4} & \textbf{39.11±0.77} & 45.57±0.76 & 43.09±0.78 & 49.97±0.86 \\
    KT~\citep{li2023knowledge}    & 34.06±0.77 & \underline{66.43±0.93} & \underline{73.10±0.87} & 22.68±0.60 & -     & -     & -     & - \\

    \midrule
    IDP (Ours) & \textbf{35.94±0.53} & \textbf{71.60±0.67} & \textbf{83.85±0.60} & \underline{23.11±0.33} & \underline{38.46±0.53} & \underline{49.70±0.61} & \textbf{44.39±0.61} & 54.07±0.63 \\
    \midrule
    \midrule
    5-shot & ISIC  & EuroSAT & CropDisease & ChestX & Car   & CUB   & Plantae & Places \\
    \midrule
    GNN~\citep{garcia2018few}   & 43.94±0.67 & 83.64±0.77 & 87.96±0.67 & 25.27±0.46 & 44.28±0.63 & 62.25±0.65 & 52.53±0.59 & 70.84±0.65 \\
    FWT~\citep{tseng2020cross}   & 43.17±0.70 & 83.01±0.79 & 87.11±0.67 & 25.18±0.45 & 44.90±0.64 & 66.98±0.68 & 53.85±0.62 & 73.94±0.67 \\
    LRP~\citep{sun2021explanation}   & 44.14±0.40 & 77.14±0.40 & 86.15±0.40 & 24.53±0.30 & 46.20±0.46 & 64.44±0.48 & 54.46±0.46 & 74.45±0.47 \\
    AFA~\citep{hu2022adversarial}   & 46.01±0.40 & 85.58±0.40 & 88.06±0.30 & 25.02±0.20 & 49.28±0.50 & 68.25±0.50 & 54.26±0.40 & 76.21±0.50 \\
    FT-All~\citep{guo2020broader} & 48.11±0.64 & 79.08±0.61 & 89.25±0.51 & 25.97±0.41 & 52.08±0.74 & 64.14±0.77 & 59.27±0.70 & 70.06±0.74 \\
    STARTUP~\citep{phoo2020self} & 47.22±0.61 & 82.29±0.60 & 93.02±0.45 & 26.94±0.94 & -     & -     & -     & - \\
    FRN$^{\dag}$~\citep{wertheimer2021few} & 47.41±0.59
 &80.77±0.60 & 91.93±0.46 & 26.77±0.40  & 49.78±0.68 & \textbf{73.06±0.72} &61.04±0.74 & 73.65±0.71 \\
    FRN$^{*}$$^{\dag}$~\citep{wertheimer2021few}  & 47.17±0.58 & 80.52±0.62 & 90.68±0.47 & 25.18±0.41 & 50.92±0.70 & 67.29±0.71 & 56.07±0.73    & 69.71±0.69 \\
    ATA-FT~\citep{wang2021cross} & 49.79±0.40 & \underline{89.64±0.30} & \underline{95.44±0.20} & 25.08±0.20 & 54.28±0.50 & 69.83±0.50 & 58.08±0.40 & \underline{76.64±0.40} \\
    NSAE~\citep{liang2021boosting}  & \underline{54.05±0.63} & 83.96±0.57 & 93.14±0.47 & \underline{27.10±0.44} & 54.91±0.74 & 68.51±0.76 & 59.55±0.74 & 71.02±0.72 \\
    BSR~\citep{liu2020feature}   & \textbf{54.42±0.66} & 80.89±0.61 & 92.17±0.45 & 26.84±0.44 & 57.49±0.72 & 69.38±0.76 & 61.07±0.76 & 71.09±0.68 \\
    ConFT~\citep{das2021importance} & 50.79±0.60 & 81.52±0.60 & 90.90±0.60 & \textbf{27.50±0.50} & \underline{61.53±0.75} & 70.53±0.75 & 62.54±0.76 & 72.09±0.68 \\
    ConFeSS~\citep{das2022confess} & 48.85±0.29 & 84.65±0.38 & 88.88±0.51 & 27.09±0.24 & -     & -     & -     & - \\
    KT~\citep{li2023knowledge}    & 46.37±0.77 & 82.53±0.66 & 89.53±0.58 & 26.79±0.61 & -     & -     & -     & - \\

    \midrule
    IDP (Ours) & 53.36±0.50 & \textbf{91.08±0.41} & \textbf{96.89±0.28} & 26.87±0.34 & \textbf{62.76±0.56} & \underline{72.92±0.58} & \textbf{69.10±0.56} & \textbf{78.08±0.55} \\
    \bottomrule
    \end{tabular}%
    }
  \label{tab:main}%
\end{table*}%

For high-order affine transformations, we resort to aligning the target and intermediate domains by Kullback-Leibler divergences. This alignment only works on the learnable scaling factor $\gamma$ and shifting factor $\beta$, fast adapting the distribution to a proper shape, as in~\figref{fig:pipeline}. We hence conduct this constraint on each target support samples $\br{x}^{ts}$ and its corresponding intermediate proxies with $\mc{R}_{\mc{P}\rightarrow \mc{T}}(\cdot)$:
\begin{equation}
\begin{aligned}
\mathcal{L}_{align}(\cdot;\beta,\gamma) &= \mathbb{D}_{K-L}(\mc{R}(\br{P},\br{V}^{t}) || \mc{R}(f_\theta(\mathbf{x}^{ts};\beta,\gamma),\br{V}^{t})) \\
                        &= \mc{R}(f_\theta(\mathbf{x}),\br{V}^{t})\log{\mc{R}(\br{P},\br{V}^{t})} + \\& 
                        \mc{R}(\br{P},\br{V}^{t})\log{\mc{R}(f_\theta(\mathbf{x}^{ts};\beta,\gamma),\br{V}^{t})}.
\label{eq:align}
\end{aligned}    
\end{equation}
where $\br{V}^{t}\in \mathbb{R}^{WH\times D \times C_{novel}}$ denotes the target learnable prototypes.

\subsection{Model Optimization}
\textbf{Learning Objective.}  Besides the pretraining stage on source domains, the target domain learning objective is composed of three constraints: 1) semantic constraints for intermediate proxies $\mc{L}_{proxy}$; 2) standard cross-entropy loss for target domain finetuning $\mc{L}_{tar}$ with $\mc{R}_{\mc{T}\rightarrow \mc{T}}(\cdot)$: 
 \begin{equation}
    \mathcal{L}_{tar}(\cdot;\theta)= 
    \mathbb{E}_{(\mathbf{x}^{ts}, \mathbf{y}^{ts}) \sim \mathcal{D}_t} 
    [\mathbf{y}^{ts}\log(\mathcal{R}(f_{\theta}(\mathbf{x}^{ts}),\br{V}^{t}))],
 \end{equation}
where $\mathbf{x}_i^{ts}$ and $\mathbf{y}_i^{ts}$ denote the target domain support set sample and its label respectively;
3) intermediate alignment loss for feature transformations $\mc{L}_{align}$. The overall learning objective has the form:
\begin{equation}
\mc{L}_{sum}=w_{t}\mc{L}_{tar}(\cdot;\theta)+ w_{p}\mc{L}_{proxy}(\cdot;\theta)+w_{a}\mc{L}_{align}(\cdot;\gamma,\beta).
\label{eq:sum}
\end{equation}
We empirically set balanced weights $w_{t,p,a}$ as 1, which already shows satisfactory performance despite its simplicity. Note that in the evaluation phase, we do not rely on the intermediate proxies and only use the backbone networks $f_{\theta}$ with target domain prototypes $\br{V}^{t}$ using measurements in Eq.~\eqref{eq:sourcerec}.  
Besides, following previous cross-domain few-shot learning methods~\citep{tseng2020cross, ijcai2021-149, hu2022adversarial}, we also incorporate the GNN model~\citep{garcia2018few} as our classifiers.  GNN learns the joint relationship of support and query samples to predict the probability of each sample belonging to each class $S$ based on the target domain reconstruction metric $\mathcal{R}(f_\theta(\mathbf{x});\br{V}^{t})$.

Alg. \ref{algo:IDP} shows the training and inference details of the proposed approach. We divide the whole learning scheme into three stages. 1) We first pre-train our model on an annotated source domain dataset, such as \textit{mini}-ImageNet \citep{vinyals2016matching}, using each pair of data $(\br{x}^s,\br{y}^s)$. 2) we fine-tuned our model using the support set of the target domain $(\br{x}^{ts},\br{y}^{ts})$ to adapt it to the target domain gradually. 3) we use the optimized model to classify the samples of the target domain query set $\br{x}^{tq},\br{y}^{tq}$.  In this manner, our training scheme shows two distinctive advantages compared to vanilla implementations: i) our approach does not rely on additional target domain unlabeled data or source domain data; ii) Our final inference network is lightweight and does not rely on additional constructed intermediate domains. These two advantages indicate our ``free lunch" implementation without any data or computation burden during the domain alignment process.

\textbf{Discussions.} The key challenge for cross-domain few-shot learning is the data scarcity and huge domain gap. Our proposed IDP benefits from two major points to solve this challenge.
\begin{enumerate}

\item Intermediate domains to reconstruct target domain content: Our approach does not directly utilize the source domain features and align this feature to the target domain, which is typically applied in the previous domain adaptation works~\citep{robey2021model, kang2018deep, zhang2019bridging}. Instead, our approach reconstructs the source domain features through an intermediate proxy, which still leans towards the target domain in terms of content. Additionally, our method allows effective control of the influence of the source domain on the intermediate domain by adjusting the source domain feature pool sizes, thereby suppressing the risk of the model overfitting during the adaptation. 

\item Optimizing normalization statistics other than CNN weights: Direct aligning with the intermediate domain might also lead to overfitting when there are only extremely few samples used for training. Thus we proposed to optimize the feature transformation statistics,~\ie, high-order and low-order statistics in the normalization layers, instead of the direct optimization on holistic network parameters. In this manner, the feature distribution can be globally transformed to align with the intermediate domain and the relative semantic relationships of the pretrained features from the source domains can be maintained, leading to the fast global alignment with extremely limited data.

\end{enumerate}

\section{Experiments}\label{sec:experiments}

\subsection{Experiment Setting}\label{sec:imp}

\textbf{Datasets.} Following the benchmark setting in CDFSL, we use the miniImageNet \citep{vinyals2016matching} training set as the source domain. Mini-ImageNet is a subset of ILSVRC-2012 \citep{deng2009imagenet}, and its training set part has 64 classes, each containing 100 natural images collected from the Internet. In addition, we use eight target domain datasets to respond to real scenarios. For diverse levels of cross-domain learning, we follow BSCD-FSL~\citep{guo2020broader} which includes CropDisease, EuroSAT, ISIC, and ChestX datasets.  For natural images in other CDFSL methods, we follow the dataset split of~\cite{tseng2020cross}, which includes Car~\citep{krause20133d}, CUB~\citep{wah2011caltech}, Plantae~\citep{van2018inaturalist} and Places~\citep{zhou2017places} datasets.

\textbf{Evaluation Protocol.} To make fair comparisons, we follow benchmark protocol~\citep{guo2020broader}, which involves validating the performance of the classifiers by simulating 600 independent 5-way $k$-shot tasks in the target domain. Since large shots can be easily learned by supervised learning, we conduct experiments with $k\in\{1, 5\}$. For each task, we randomly select 5 categories from all categories in the target domain dataset and, within each category, we randomly select $k$ images for the support set and 16 images for the query set. For each task, we fine-tune the pre-trained model on the support set and evaluate its performance on the query set. We repeat this process 600 times for each experiment setting, resulting in 600 fine-tuned and evaluated models. Average classification accuracy and its 95\% confidence interval on the query set are reported in accordance with the benchmark evaluations.

\begin{table*}[!t]
    \caption{Comparisons with state-of-the-art models using the 5-way random-shot setting on Meta-Dataset benchmark. The performances are evaluated using official codes and the best values on each set are highlighted in bold.}
    \centering
    \begin{tabular}{ccccc}
    \hline
        Test Dataset & KT~\cite{li2023knowledge} & LDP-Net~\cite{zhou2023revisiting} & TSA~\cite{li2022cross} & IDP(ours) \\ \hline
        ILSVRC & 62.51 & 59.86 & \textbf{80.37} & 77.42 \\ 
        Omniglot & 79.34 & 90.15 & 93.01 & \textbf{98.64} \\ 
        Aircraft & 44.48 & 60.94 & 61.88 & \textbf{75.14} \\ 
        Birds & 55.86 & 57.51 & \textbf{84.80} & 71.28 \\ 
        Textures & 65.59 & 65.35 & 75.76 & \textbf{78.82} \\ 
        Quick Draw & 71.97 & 83.98 & 78.71 & \textbf{87.64} \\ 
        Fungi & 50.82 & 45.84 & \textbf{69.98} & 68.46 \\ 
        VGG Flower & 78.84 & 82.83 & 91.63 & \textbf{96.28} \\ 
        Traffic Sign & 56.42 & 86.04 & 75.07 & \textbf{88.04} \\ 
        MSCOCO & 58.71 & 62.17 & 72.90 & \textbf{76.16} \\ \hline
        Avg. & 62.45 & 69.47 & 78.41 & \textbf{81.79} \\ \hline
    \end{tabular}
    \label{tab:meta}
\end{table*}

\textbf{Implementation Details.} To make a fair comparison with existing works \citep{guo2020broader, phoo2020self, liang2021boosting}, in all experiments we use ResNet-10 backbone network with SGD optimizer. For pretraining, we set the learning rate to 0.05 and the batch size to 64 for 350 epochs. We then conduct meta-finetuning on each target domain for 50 epochs with a learning rate of 0.01. We found that setting the prototype number of each class $\br{V}_i^{t}$ to 20 is sufficient to obtain satisfactory results. To ensure that each pixel on the feature map has a sufficient field of perception, we set $W=H=5$, thus the input image resolution is scaled to $160\times 160$.  Following previous works~\citep{tseng2020cross, ijcai2021-149, hu2022adversarial}, we also used a meta-trained lightweight GNN~\citep{garcia2018few} as final classifiers to learn the relationship of few-shot samples to obtain the final results. As the test phase required less graphics memory and could be executed on a lower-performance GPU, our model was implemented in the PyTorch \citep{paszke2019pytorch} framework on a single NVIDIA GTX3090 GPU.

\subsection{Comparison with State-of-The-Art}
\textbf{Benchmarking on Domain with Diverse Levels.} We first conduct comparisons on the most widely-used BSCD-FSL benchmark \citep{guo2020broader} with the state-of-the-art models. Tab. \ref{tab:main} exhibits the results of different levels of domain transfer,~\ie, a gradual decrease of visual features from CropDisease and EuroSAT to ISIC and ChestX shared with the source domain.  For fair comparisons, we also extend the FRN~\citep{wertheimer2021few} on this cross-domain few-shot setting with the prototypical networks~\citep{snell2017prototypical} on cross-domain classes and we finetuned FRN with the identical hyper-parameters and optimizer, noted as FRN$^{*}$. Identical to our method, FRN$^{*}$ adopts the SGD optimizer with a learning rate of 0.01 and performs meta-finetuning on each target domain for 50 epochs.
From Tab.~\ref{tab:main}, the finetuning on FRN leads to an inferior performance than FRN, indicating that the FRN representations are easy to overfit on few-shot samples, especially on these datasets with huge domain gaps. Note that FRN methods are built upon the ResNet-12 backbone while others are using the lightweight ResNet-10 backbones. Our method on average achieves 53.63\% and 67.05\% on 5-way 1-shot and 5-shot settings respectively, outperforming all the listed CDFSL competitors significantly, including  ConFT~\citep{das2021importance}, ConFeSS~\citep{das2022confess}, and KT~\citep{li2023knowledge}, and our proposed IDP leads a new state-of-the-art. 

\begin{table}[t]
    \centering
    \caption{Ablation studies of 5-way K-shot learning of different modules.}

    \resizebox{\linewidth}{!}{
    \begin{tabular}{ccc|cc}
    \toprule
    \multirow{2}[4]{*}{Method} & \multicolumn{2}{c|}{EuroSAT} & \multicolumn{2}{c}{Places} \\
\cmidrule{2-5}          & 1-shot & 5-shot & 1-shot & 5-shot \\
    \midrule
    GNN (base) & 63.69±1.0 & 83.64±0.8 & 53.10±0.8 & 70.84±0.7 \\
    +$\mc{L}_{tar}$ & 66.82±0.6 & 87.59±0.4 & 53.39±0.7 & 74.34±0.6 \\
    +$\mc{L}_{proxy}$   & 68.95±0.7 & 90.11±0.5 & 53.73±0.7 & 77.37±0.6 \\
    +$\mc{L}_{align}$ & \textbf{71.60±0.7} & \textbf{91.08±0.4} & \textbf{54.07±0.6} & \textbf{78.08±0.6} \\
    \bottomrule
    \end{tabular}%
    }
    \label{tab:ablation}
\end{table}

\begin{table}[t]
    \centering
    \caption{Effects of different optimization combinations on 5-way K-shot learning performance. $\bs{BN}_{l}$ and $\bs{BN}_{h}$ indicates the low-order and high-order statistics.}
    \setlength{\tabcolsep}{1.0mm}
    \renewcommand{\arraystretch}{1.3}
    \resizebox{\linewidth}{!}{
    \begin{tabular}{ccccc|cc}
    \toprule
    \multicolumn{3}{c}{Optim.} & \multicolumn{2}{c|}{EuroSAT} & \multicolumn{2}{c}{Places} \\
\cmidrule{4-7}    $f_{\theta}$ & $\bs{BN}_{l}$ & $\bs{BN}_{h}$ & 1-shot & 5-shot & 1-shot & 5-shot \\
    \midrule
         &      &      & 69.54±0.7 & 88.74±0.4 & 52.30±0.7 & 76.83±0.6 \\
         &      & \checkmark     & 70.66±0.7 & 90.76±0.4 & 53.83±0.7 & 77.78±0.6 \\
    \checkmark     & \checkmark     & \checkmark     & 69.18±0.7 & 88.84±0.5 & 53.72±0.7 & 77.32±0.6 \\
         & \checkmark     & \checkmark     & \textbf{71.60±0.7} & \textbf{91.08±0.4} & \textbf{54.07±0.6} & \textbf{78.08±0.6} \\
    \bottomrule
    \end{tabular}%
    }
    \label{tab:optim}
\end{table}

\begin{figure*}[t]
\centering
\includegraphics[width=\textwidth]{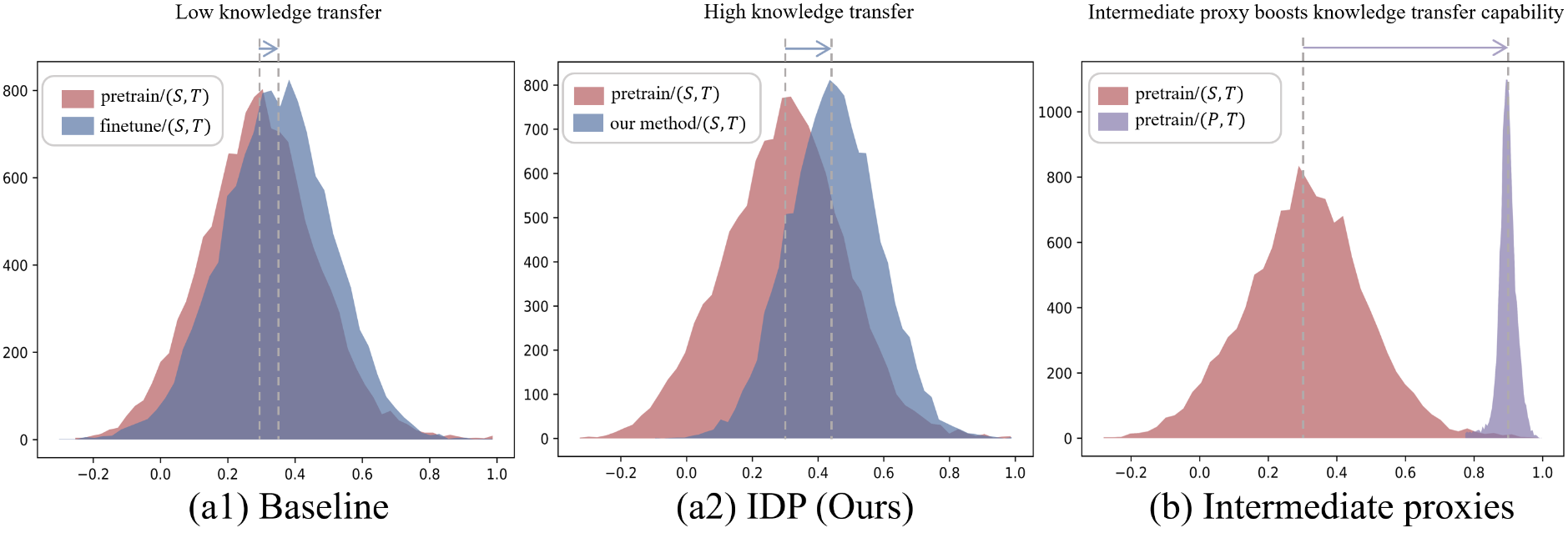}
\caption{  Frequency distribution of randomly sampled distances between feature pairs from different domains (\textit{Higher values indicates better domain alignments.}).  a1) and a2): Domain-transferring ability of \textbf{Baseline} and \textbf{Ours}. b):  \textbf{Intermediate domain proxies} boost knowledge-transferring capabilities. }
\label{fig:distrib}
\end{figure*}

\textbf{Benchmarking on Natural Images.} Besides the diverse domain benchmarking, the other line of work focuses on natural images but with different data distributions. Tab. \ref{tab:main} presents the comparisons on natural images,~\eg, cars and birds, with representative state-of-the-art methods including NSAE~\citep{liang2021boosting}, ConFT~\citep{das2022confess}. Similar to the CDFSL benchmark, following the class split of~\cite{tseng2020cross}, we also conduct experiments on FRN and finetuned FRN (FRN$^{*}$) on these natural domain images.Note that FRN~\citep{wertheimer2021few} adopted ResNet-12 as backbones while others used ResNet-10. FRN shows better performance in a 1-shot setting when there are fewer domain gaps,~\eg, CUB datasets. When more samples are available (\ie, 5-shot), our method surpasses FRN by over 6.33\% on average. Besides, it can be observed that our method shows leading results to prevailing methods under this natural domain setting. Among them, our method outperforms the second-place method ConFT~\citep{das2022confess} by over 4\% on average for the 5-way 5-shot setting, indicating the strong generalization capabilities of our method.

\textbf{Extensions on Meta-Dataset.} 
Compared to the prevailing datasets, the Meta-Dataset~\citep{triantafillou2019meta} is a large-scale benchmark consisting of ILSVRC-2012 \citep{deng2009imagenet} as the training set and ten individual datasets as the testing set. Additionally, the Meta-Dataset benchmark~\citep{triantafillou2019meta} introduces varying samples for each class, determined based on the distribution of real-world data. For each task, we randomly select 5 categories from all categories in the test dataset. Following~\cite{triantafillou2019meta}, the sample size for each category's support set is a random number in $[1,100]$, while the number of query samples is fixed at 10, as all categories hold equal importance. We computed the average accuracy and 95\% confidence interval for 600 independently sampled tasks while keeping the other implementation details consistent with the CDFSL benchmark. The results of our method are presented in~\tabref{tab:meta}. We select three state-of-the-art methods for comparison,~\ie, KT~\citep{li2023knowledge}, LDP-Net~\citep{zhou2023revisiting}, and TSA~\citep{li2022cross}. We conduct the 5-way random-shot experiments with their open-source code for fair comparisons. Compared with these methods, our IDP maintains its leading position on the Meta-Dataset benchmark and achieves clear improvements on multiple subsets, indicating its powerful generalization ability to adapt to real-world scenarios.

\begin{figure}[t]
    \centering
    \includegraphics[width = 0.5\textwidth]{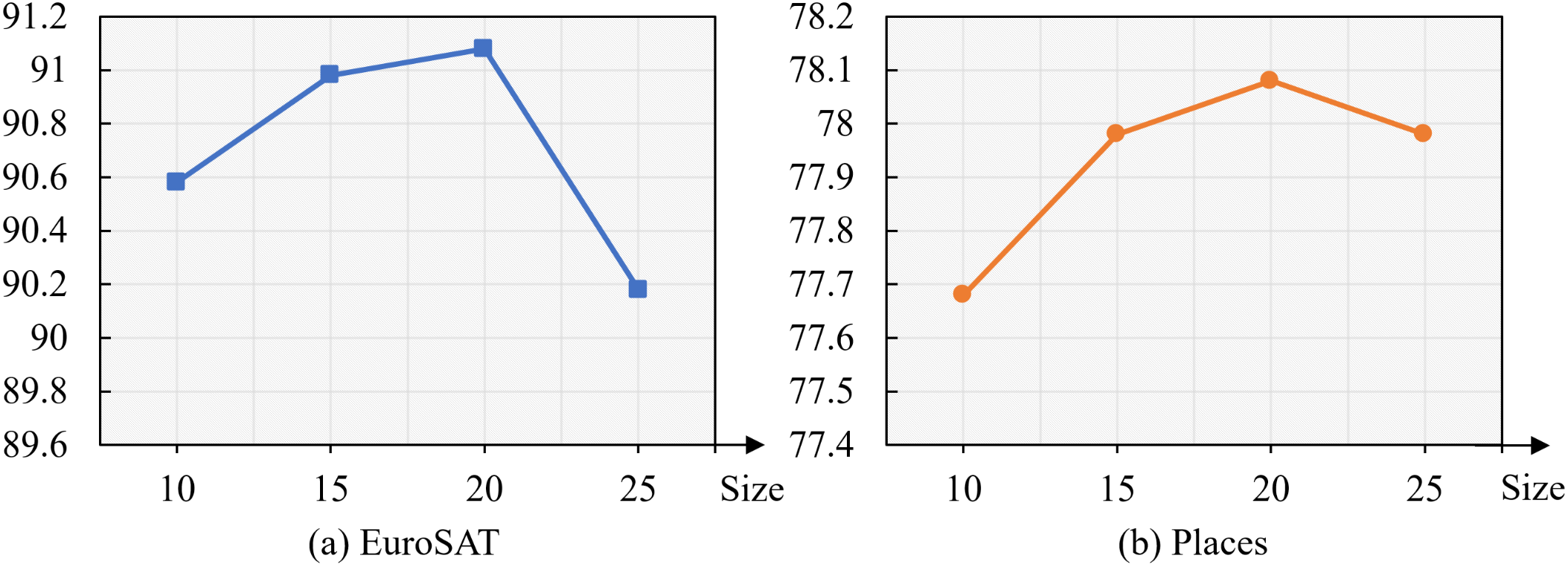}
    \caption{The effect of different size of class prototype $\br{V}_i^{t}$. All experiments are conducted under 5-way 5-shot conditions, and the vertical coordinates indicate the performance of our method.}
    \label{fig:hyper_proto}
\end{figure}

\begin{figure}[t]
    \centering
    \includegraphics[width = 0.5\textwidth]{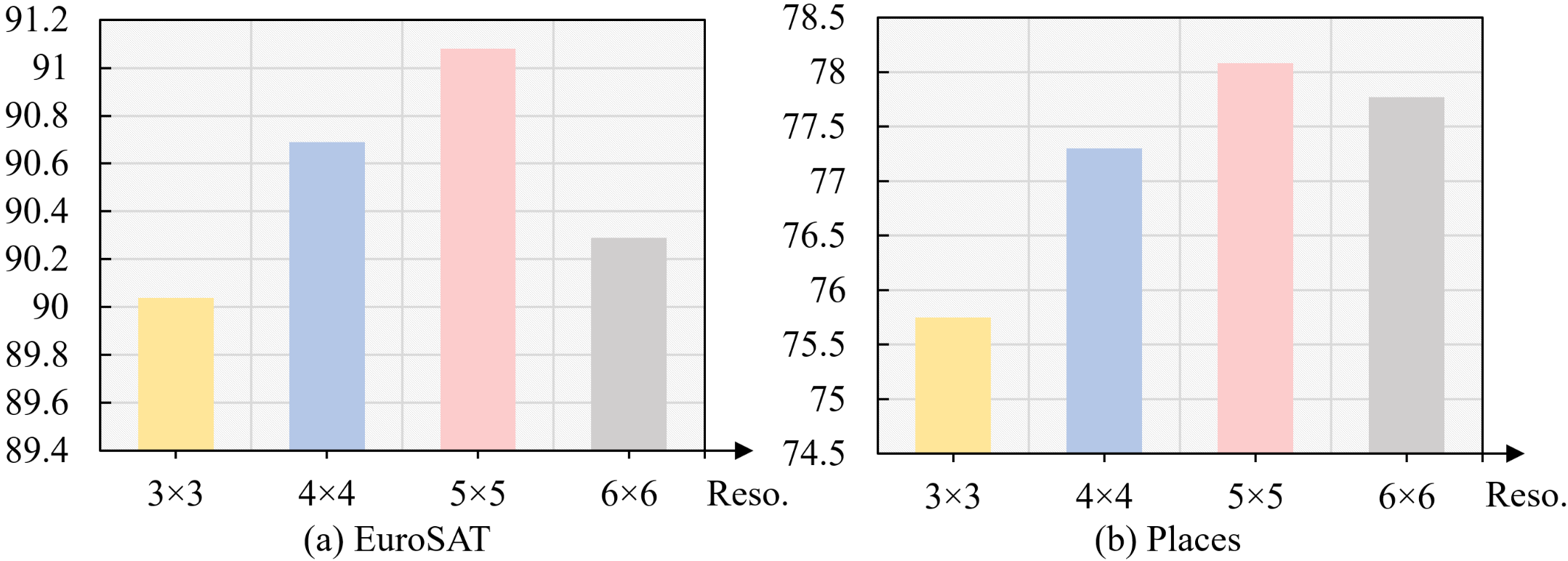}
    \caption{The effect of resolution of the feature map. All experiments were conducted under 5-way 5-shot conditions, and the vertical coordinates indicate the performance of our method.}
    \label{fig:hyper_reso}
\end{figure}

\subsection{Performance Analyses}

\textbf{Effect of Different Components.} To evaluate the effectiveness of our method, we perform ablation studies in Tab.~\ref{tab:ablation}. In the first row, we show the performance of the baseline model GNN, $\mc{L}_{tar}$ represents the addition measurements to obtain the predicted values. It can be found that this metric effectively improves the discriminative ability of the model. Further, we use the spatial feature pool $\mc{U}$ to reconstruct the target domain support set samples to obtain intermediate proxies $\br{P}$. After that, we optimize the classifier with the cross-entropy loss $\mc{L}_{proxy}$ of these intermediate proxies with labels as shown in the third row, which improves the cross-domain generalizability of the model. We eventually add intermediate proxy alignment loss $\mc{L}_{align}$ to further improve the final performance.

\begin{figure*}[t]
    \centering
    \begin{minipage}[!t]{0.65\textwidth}
        \centering
        \includegraphics[width=\textwidth]{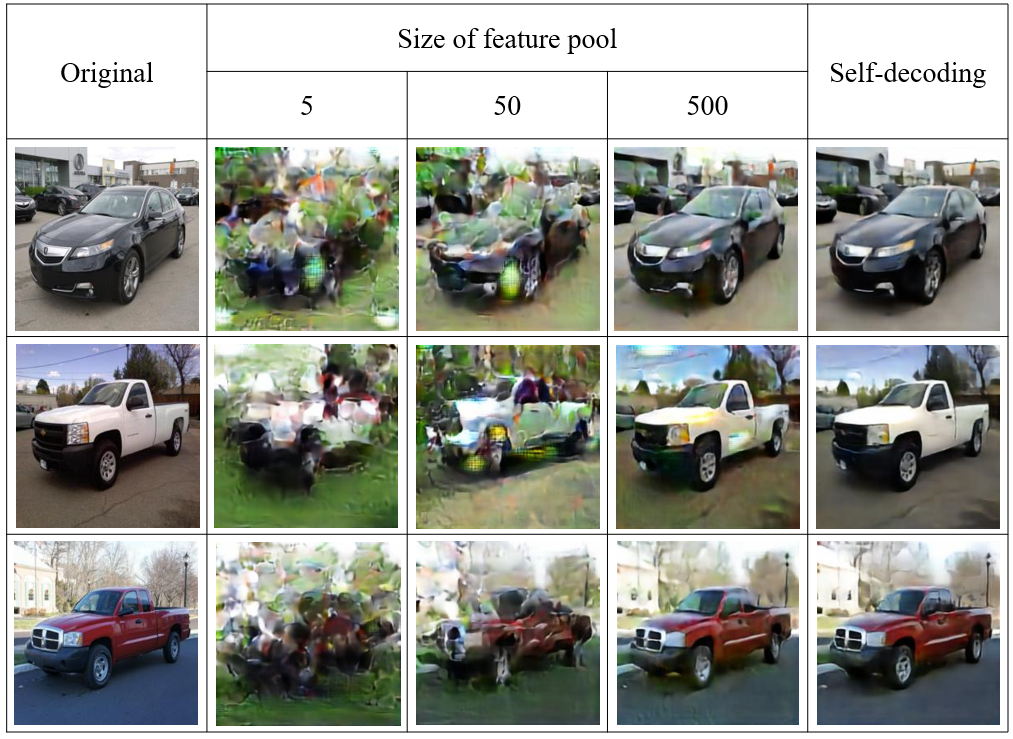}
        \caption{Visualization of reconstruction using different source domain feature pool sizes.}
        \label{fig:size}
    \end{minipage}
    \hfill
    \begin{minipage}[!t]{0.28\textwidth}
        \centering
        \includegraphics[width=\textwidth]{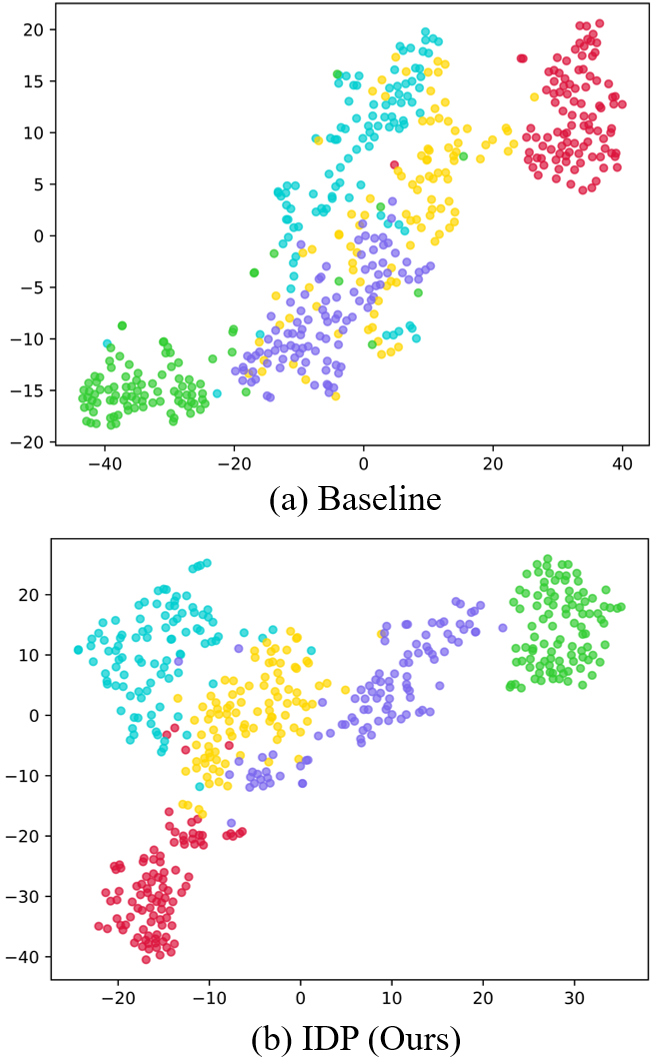}
        \caption{t-SNE comparisons of 5-way classification on target domains (EuroSAT).}
        \label{fig:tsne}
    \end{minipage}
\end{figure*}

\textbf{Variant of Alignment Loss Optimization for Intermediate Proxies.} In Tab. \ref{tab:optim}, we compare the effects of optimizing different components of the network. where $f_{\theta}$ represents the optimization of the whole network parameters, $\bs{BN}_{l}$ represents the tuning of the statistics of the BN layer using intermediate proxies, and $\bs{BN}_{h}$ represents the optimization of the learnable scaling and shifting parameters of the BN layer. Comparing the first two rows we can observe that the proposed strategy for adjusting the BN layer statistics is effective. Further, we find that optimizing the entire network parameters using alignment loss $\mc{L}_{align}$ in the third row overfits the model to the intermediate proxies, which impairs the performance. Finally, we use alignment loss to optimize only the higher-order learnable parameters of the BN layer $\bs{BN}_{h}$, achieving the highest performance in the last row.

\textbf{Effect of Intermediate Proxy Alignment Loss on Distribution.} In Fig. \ref{fig:distrib}, we show the difference between the distribution of the source and target domains (a1) without alignment loss $\mc{L}_{align}$ and (a2) with alignment loss. We conduct the following steps to calculate the alignment scores: a) Randomly sampling pairs of data from different domains; b) Calculating the Euclidean distance of their L2 normalized representations; c) Plotting the frequency distribution of distances within small intervals. It can be observed that the addition of the alignment loss significantly reduces the source and target domain metrics,\ie, the difference distribution is closer to 1. This shows that our approach can better transfer the knowledge learned by the model during the pre-training phase on the source domain to the target domain. \ref{fig:distrib}(b) shows the difference between the distribution of target domain samples and intermediate proxies, and it can be seen that the intermediate proxies are closer to the target domain, hence it is easier to align them.

\textbf{Effect of Prototype Size.} In Fig. \ref{fig:hyper_proto}, we exhibit how different sizes of class prototype $\br{V}_i^{t}$ affect the performance of our method. It can be observed that on the one hand if the prototype size is too small then the representation of the categories is insufficient, on the other hand, a large prototype size reduces the discriminative ability of the model.

\textbf{Effect of Feature Resolution for Reconstruction.}  In Fig. \ref{fig:hyper_reso}, we present how the resolution of the feature map affects the performance of our method. It can be observed that the size of the image region corresponding to each vector on the feature map affects the retention of details by the network. Smaller scales ($3 \times 3$) of the network features usually lead to the coarse observation of local details, while larger scales ($7 \times 7$) lead to an inferior focus on the global reception field.

\textbf{Extensions on Intra-domain Gap.} To evaluate the effectiveness of our method, the extreme scenarios for cross-domain learning is ``intra-domain",~\ie, evaluated on the same miniImageNet dataset~\citep{vinyals2016matching}. We compare with the recent few-shot learning method on the mini-ImageNet dataset. Our proposed method uses the ResNet-10 network as backbones but still achieves good performance in this intra-domain setting, as in~\tabref{tab:intradomain}, which indicates the strong generalization ability when facing fewer domain gaps. 

  \begin{table}[!t]
    \caption{Comparisons with state-of-the-art models on mini-ImageNet benchmark dataset. The best values on each set are highlighted in bold. $\dag$: using ResNet-12 backbones. $*$: using lightweight ResNet-10 backbone.}

    \centering
    \resizebox{\columnwidth}{!}{
    \setlength{\tabcolsep}{1.0mm}
    \renewcommand{\arraystretch}{1.3}
    \begin{tabular}{ccc}
    \hline
        Method & 5-way 1-shot & 5-way 5-shot \\ \hline
         FEAT$\dag$~\citep{ye2020few} & 66.78±0.20 & 82.05±0.14 \\ 
         H-OT$\dag$~\citep{guo2022adaptive}& 65.63±0.32 & 82.87±0.43 \\ 
         SAPENet$\dag$~\citep{huang2023sapenet}& 66.41±0.20 & 82.76±0.14 \\
         \hline
        MatchingNet$*$~\citep{vinyals2016matching} & 58.76±0.61 & 72.53±0.69 \\ 
         RelationNet$*$~\citep{sung2018learning}& 58.64±0.85 & 73.78±0.64 \\ 
         GNN$*$~\citep{garcia2018few}& 66.32±0.80 & 81.98±0.55 \\ 
         IDP(Ours)$*$ & \textbf{67.16±0.71} & \textbf{84.64±0.46} \\ \hline
    \end{tabular}
    }
    \label{tab:intradomain}
\end{table}

\textbf{Computational Efficiency.} Our proposed method shows similar computational costs compared to the baseline methods~\citep{garcia2018few}. Benefiting from the optimization scheme, the intermediate domain proxies are dropped during the inference stage, and our method does not rely on many additional network parameters. The detailed inference time and computational costs are presented in~\tabref{tab:computation}. With the additional 8.5\% costs, our method improves the baseline by a large margin,~\eg, 43.96\% to 53.36\% on the ISIC dataset.

 \begin{table}[!t]
    \centering
    \caption{Comparison of computational efficiency with representative methods.}
    \resizebox{\columnwidth}{!}{
    \setlength{\tabcolsep}{1.0mm}
    \renewcommand{\arraystretch}{1.3}
    \begin{tabular}{ccc}
    \hline
        Method & GFLOPS & Time (ms) \\ \hline
        GNN~\citep{garcia2018few} & 189.0 & 14.6 \\ 
        GNN-ATA~\citep{wang2021cross} & 196.1 & 15.9 \\ 
        TPN-ATA~\citep{wang2021cross} & 211.6 & 17.3 \\ 
        KT~\citep{li2023knowledge} & 214.1 & 19.6 \\ 
        IDP (Ours) & 205.2 & 19.3 \\ \hline   
    \end{tabular}
    }
    \label{tab:computation}
\end{table}

\subsection{Visualization and Explanations}

\textbf{Effect of Source Domain Feature Pool.}
In Fig. \ref{fig:size}, we show the relationship between the number of feature pool sizes and the reconstruction. It can be observed that as the pool of features involved in the reconstruction becomes larger, the intermediate proxies are able to reconstruct the target domain samples more clearly. Conversely, when the proxy pool is relatively small ($\leq50$), the reconstructed intermediate proxies are more ambiguous and show more of the source domain style. This indicates that the intermediate domain reconstruction is \textit{controllable} when changing the reconstruction materials in the stored feature pool. 

\textbf{Target Domain Category Embedding.}
We visualize the target domain embeddings in the EuroSAT dataset in Fig. \ref{fig:tsne} using t-SNE. Fig. \ref{fig:tsne} shows that our methods generate few intra-class differences and larger inter-class differences compared to the baseline. This indicates after domain adaptation, our proposed method retains a stronger discriminative capability with only a few given samples.

\section{Conclusions and Limitations}\label{sec:conclusion}
In this paper, we start from a different view to revisit the problem of cross-domain few-shot learning. Prevailing research efforts mainly focus on the generalized representation of feature learning while neglecting the fast domain alignment with these few samples. Toward this end, we propose to reconstruct an intermediate domain using source embeddings and use the reconstructed domain proxies to develop a fast domain transformation technique with normalization layers. Despite its superior performance on public CDFSL benchmarks, our proposed method still relies on dense feature reconstructions, which may limit the extension of our work on segmentation and dense estimation vision tasks, which we leave for our future exploration.

\begin{acknowledgements}
This work is partially supported by grants from the National Natural Science Foundation of China under contracts No. 62132002, No. 62202010, and in part by the Fundamental Research Funds for the Central Universities.
\end{acknowledgements}

\section*{Appendix}

\subsection*{A. Additional Details of the Empirical Study}\label{app:A}

\subsection*{A.1 Implementation Details}

\begin{figure*}[htbp]
    \centering
    \includegraphics[width = 0.875\textwidth]{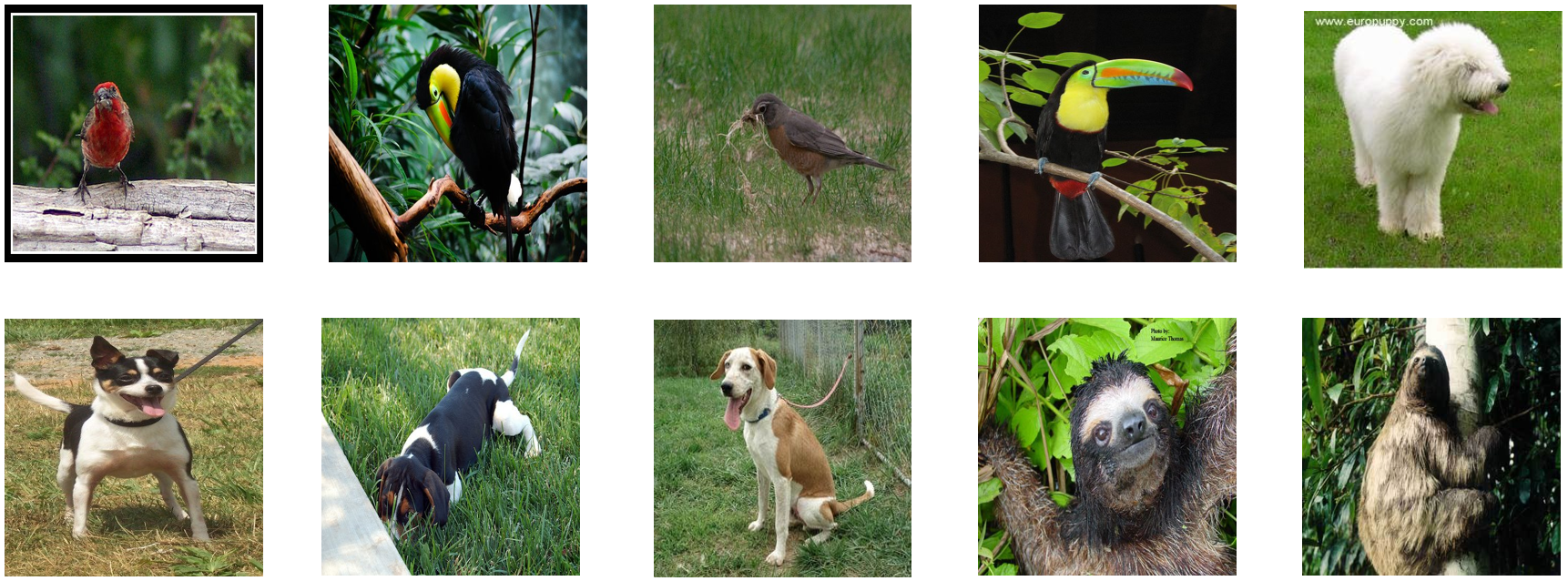}
    \caption{More illustration of objects in domain $\mc{A}$. Domain $\mc{A}$ consists of partial \textit{mini}-ImageNet dataset categories, including birds, dogs, and sloths, which are objects in the jungle or on the grass.}
    \label{fig:domain_a}
\end{figure*}

\begin{figure*}[htbp]
    \centering
    \includegraphics[width = 0.875\textwidth]{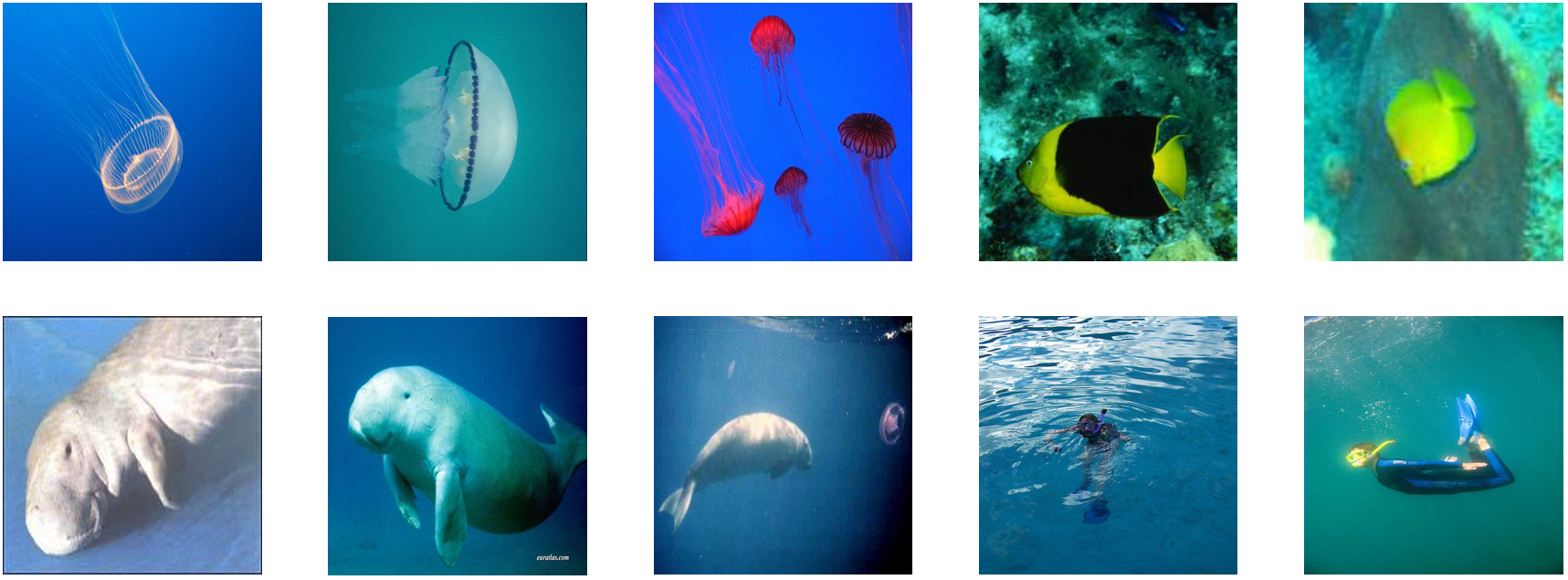}
    \caption{More illustration of objects in domain $\mc{B}$. Domain $\mc{B}$ consists of partial \textit{mini}-ImageNet dataset categories, including jellyfish, manatee, and butterfly fish, which are underwater objects.}
    \label{fig:domain_a2}
\end{figure*}

To conduct the empirical study, we first organize two ``sub-Domains", domain $\mc{A}$ and domain $\mc{B}$. Both ``sub-Domain" are sampled from \textit{mini}-ImageNet \citep{vinyals2016matching} and contain a series of stylistically distinct and visually similar image classes. We give more examples of domain $\mc{A}$ and domain $\mc{B}$ in Fig. \ref{fig:domain_a} and \ref{fig:domain_a2}, where it can be observed that since objects in domain $\mc{A}$ are often located in the jungle or on grass, they are visually greenish in color; in contrast, objects in domain $\mc{B}$ are often located underwater and are visually bluish in color. The impact of feature base $\{\br{C}_i\}_{i=1}^n$ on domain reconstruction can be inferred by observing the performance of the reconstructed intermediate domain proxies $\mc{P}$ in terms of style and content.

For image level reconstruction $\bs{img} \xrightarrow{} \bs{r.img}$
, we first resize the image to $224\times224$ and slice the image into blocks of size $7\times7$, each with a length and width of $32$. feature bases $\{\br{C}_i\}_{i=1}^n$. We flatten these image blocks into vectors, which are used as our feature bases $\{\br{C}_i\}_{i=1}^n \in \mathbbm{R}^{n\times 1024}$. It is worth noting that since the reconstruction process is independent of the spatial location of the pixels, the flattening operation does not affect the results of the image block reconstruction. We then solve for the intermediate agent $\mc{P}$ according to Eq. \ref{eq:solution}.

For feature level reconstruction $\bs{feat} \xrightarrow{} \bs{r.feat}$
, we still resize the image to $224\time 224$, which will result in an output feature map size of $7\times 7$ for backbone network $f(\theta)$. We use the pixels on the feature map as feature bases $\{\br{C}_i\}_{i=1}^n \in \mathbbm{R}^{n\times d}$ and perform a reconstruction process similar to the image level reconstruction described above, where $d$ is the output feature dimension of the $f(\theta)$.

\subsection*{A.2 Visualization Method}

For image level reconstruction $\bs{img} \xrightarrow{} \bs{r.img}$, we reshape the reconstructed vectors into image blocks and stitch these image blocks into a new image according to their spatial locations. For feature level reconstruction $\bs{feat} \xrightarrow{} \bs{r.feat}$, we obtain feature maps as the reconstruction results. However, since these feature maps cannot be directly visualized, we propose to utilize a decoder to convert them into images. Table \ref{tab:decoder} illustrates the architectural specifications of the decoder, which can be seen as a mirrored version of the ResNet10 backbone network. It consists of Deconv blocks, each Deconv block containing an upsampling function, a convolution operator, and a ReLU activation. We train the decoder to decode the original images from the feature maps, which are generated by the encoder. Finally, we utilize the decoder to visualize the intermediate representations.

\subsection*{A.3 More Visualization Results}

\begin{figure*}[htbp]
    \centering
    \includegraphics[width = 0.875\textwidth]{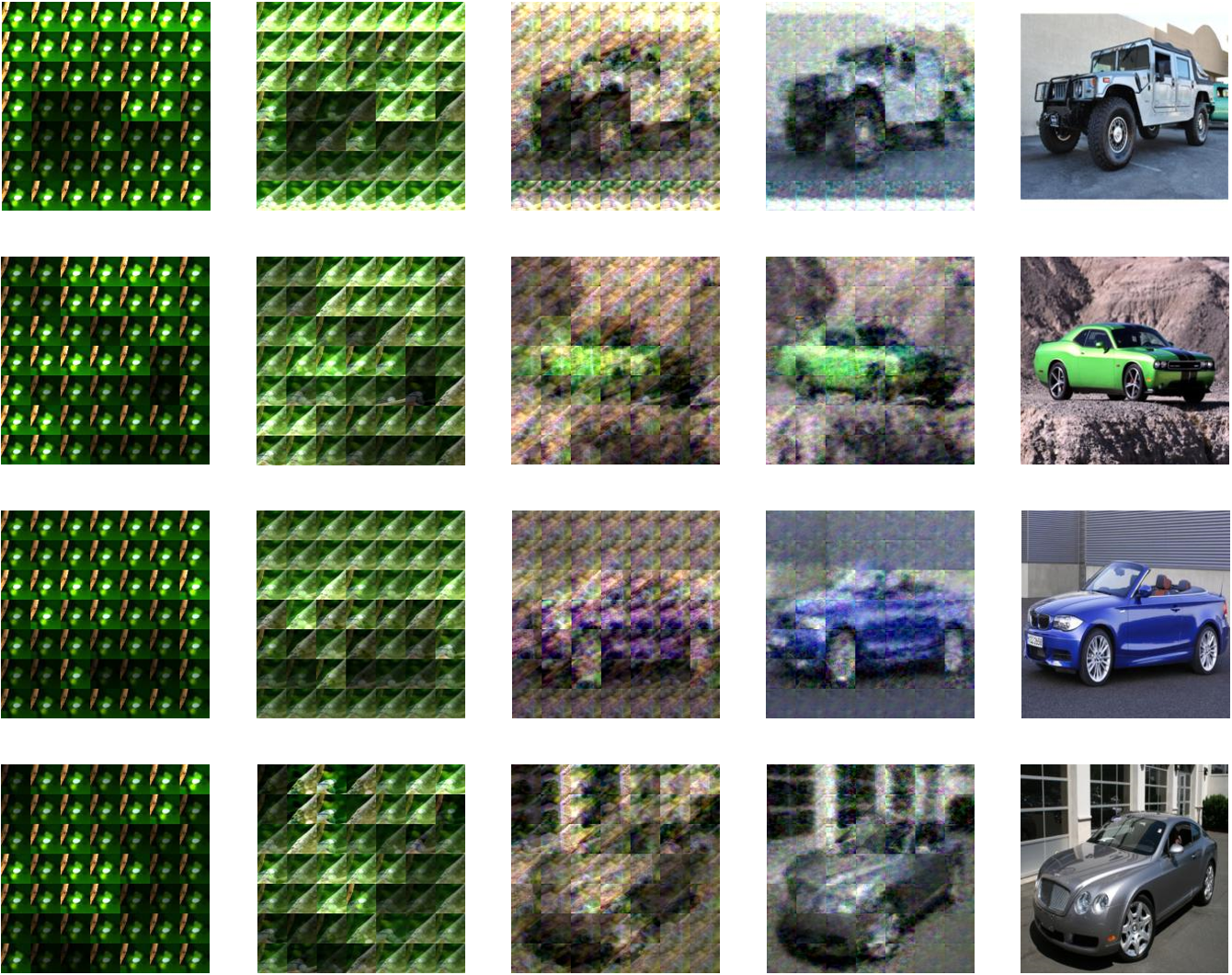}
    \caption{Illustration of image level reconstruction $\bs{img} \xrightarrow{} \bs{r.img}$. The source domain is set to domain $\mc{A}$. The number of image blocks involved in the reconstruction from left to right in the first four columns is $n=\{1,5,20,50\}$, respectively. The rightmost column is the target image.}
    \label{fig:img_recon}
\end{figure*}

In Fig. \ref{fig:img_recon}, we visualize more intermediate proxies for image-level reconstructions. We can observe a strong relationship between the number of image blocks involved in the reconstruction and the reconstruction results. When reconstructing the target image using only one image block,~\ie, the first column of Fig. \ref{fig:img_recon}, the intermediate proxy is simply a concatenation of different brightness and contrast combinations of that image block. As the number of image blocks involved in the reconstruction increases, the intermediate agents behave from blurred to clear and eventually very close to the target image.

\begin{table}[t]
  \centering
  \caption{The architecture specifications of the decoder modules in ResNet10. We insert a BatchNorm layer behind each Deconv layer.}
  \setlength{\tabcolsep}{1.0mm}
\renewcommand{\arraystretch}{1.0}
    \begin{tabular}{c|c}
    \toprule
    \textbf{Module} & \textbf{Specifications} \\
    \midrule
    \multirow{5}[2]{*}{ResNet10}
          & 3×3 Deconv-ReLU, 256 filters, stride 2, padding 1 \\
          & 3×3 Deconv-ReLU, 128 filters, stride 2, padding 1 \\
          & 3×3 Deconv-ReLU, \ \,64 filters, stride 2, padding 1 \\
          & 3×3 Deconv-ReLU, \ \,32 filters, stride 2, padding 1 \\
          & 3×3 Deconv-Sigmoid, 3 filters, stride 2, padding 1 \\
    \bottomrule
    \end{tabular}%
  \label{tab:decoder}%
\end{table}%

\subsection*{B. Proof of Proposition 1}

\textbf{Proposition 1}\textit{ (High semantic similarity).}
\textit{By controlling the ridge regression regular term $\lambda$ , the semantic similarity between the intermediary domain proxy $\mathcal{P_\lambda}$ and the target domain $\mathcal{T}$ is larger than that between the source domain $\mathcal{S}$ and the target domain $\mathcal{T}$. Their inter-domain discrepancy distance satisfies the following relationship:}
$\exists~\lambda$ \textit{,s.t.} $disc_\mathcal{L}(\mathcal{S}, \mathcal{T}) > disc_\mathcal{L}(\mathcal{P}, \mathcal{T})$

\textit{Proof.} We first recall the formula for ridge regression:
\begin{equation}
  \widehat{\br{W}} = \arg{\min\limits_{\br{W}}\left\| {\br{T} - \br{W}\br{U}} \right\|^2} + \lambda\left\| \br{W} \right\|^2, \tag{16}
\end{equation}

where $\br{U} \in \mathbb{R}^{n\times d}$ denote the generalized representation of each source domain visual pattern and $\br{T}\in \mathbb{R}^{r\times d}$ denote the target domain embeddings required to be reconstructed. The hyper-parameter $\lambda$ is used to balance the regularization of the $\ell_2$-norm. As stated in Section \ref{sec:domain_alignment}, we begin by resizing the feature pool through clustering mapping $\mc{M}: \mathbb{R}^{n\times d}\rightarrow \mathbb{R}^{r\times d}$. This operation aligns the size of $\br{U}$ with $\br{T}$ and sets the dimension of $\br{W}$ to $r\times r$.

Next, we define $\br{F}(\lambda) = \|\br{T}-\br{P}\|^2 - \|\br{T}-\br{U}\|^2$ as the difference from the $\br{T}$ to the $\br{P}$ and from $\br{T}$ to $\br{U}$. Since $\widehat{\br{W}}=\br{T}\br{U}^{\top}(\br{U}\br{U}^{\top}+\lambda\br{I})^{-1}$ is the ridge regression closed-form solution, \br{F} can be rewritten as:
\begin{equation}
  \br{F}(\lambda) = \|\br{T}-\br{T}\br{U}^\top (\br{U}\br{U}^\top +r\br{I})^{-1}\|^2 - \| \br{T}-\br{U} \|^2. \tag{17}
\end{equation}
We then solve for the partial derivative of $\br{F}$ with respect to $r$.
\begin{equation}
  \frac{\partial \br{F}(\lambda)}{\partial r} = 2\mathrm{tr}(\br{H}(\br{T}^\top -\br{H}\br{U}\br{T}^\top)\br{T}\br{U}^\top \br{H}), \tag{18}
\end{equation}
where $\mathrm{tr}$ denotes the trace of the matrix and $\br{H} = (\br{U}\br{U}^\top +r\br{I})^{-1}$ in order to simplify the expression. It can be observed that the monotonicity of $\br{F}$ is determined by both the regular term $r$ and the distribution of the data. Therefore, we discuss it by situation:

\textbf{If $\bm{\lambda} $ is equal to 0.}
According to the property of ridge regression, we can obtain that for any $\br{W}^{*}\in \mathbbm{R}^{n \times n}$, the target domain embeddings $\br{T}$ satisfies 
\begin{equation}
\left \| \br{T}-\widehat{\br{W}}\br{U} \right \| ^2 + \lambda \left \| \widehat{\br{W}} \right \| ^ 2 \leq \left \| \br{T}_i-\br{W}^{*}\br{U} \right \| ^2 + \lambda \left \| \br{W}^{*} \right \|  ^ 2, \tag{19}
\end{equation}
where $\widehat{\br{W}}=\br{T}\br{U}^{\top}(\br{U}\br{U}^{\top}+\lambda\br{I})^{-1}$ is the ridge regression closed-form solution. Assuming $\br{T}\neq \br{U}$, we then substitute the identity matrix $\br{I}\in \{0, 1\}^{r \times r}$ for $\br{W}^{*}$ in this equation, and obtain:
\begin{equation}
\left \| \br{T}-\widehat{\br{W}}\br{U} \right \| ^2 + \lambda \left \| \widehat{\br{W}} \right  \| ^ 2 < \left \| \br{T}-\br{U} \right \| ^2 + \lambda \left \| \br{I} \right  \| ^ 2. \tag{20}
\label{eq:ineq}
\end{equation}
Bringing $\lambda=0$ into the Eq. \eqref{eq:ineq}, we can prove that $\br{F}<0$.

\textbf{If $\bm{\lambda > 0}$ and F is monotonically increasing}. 
Data noise may causes $\left \| \br{T}-\br{W}\br{U} \right \| ^2$ to be monotonically increasing, since $r=0$ with $\br{F}(r)<0$ holds, there must exist a $\lambda=\lambda'$ near 0 for $\br{F}(\lambda)<0$ to hold.

\textbf{If $\bm{\lambda > 0}$ and F is decreases first}
, there will exist $\lambda>\lambda'$ such that $\br{F}(\lambda)<0$. 

According to Definition 1, we can conclude that Proposition 1 holds. $\hfill \square$

\subsection*{C. Proof of Proposition 2}
\textbf{Proposition 2} \textit{(Reducing target classification error).}
\textit{Aligning the target domain $\mc{T}$ to the intermediate domain proxy $\mc{P}_\lambda$ can reduce the discrepancy distance between the source and target domain $disc_\mc{L}(\mc{S}, \mc{T})$, which in turn reduces the error of the classifier $\epsilon_\mc{T}$ on the target domain.}

\textit{Proof.} We start by defining the symbols. Specifically, we denote $l$ as the class labeling function, with $l_\mc{S}$ representing the function for the source domain and $l_{T}$ representing the function for the target domain. 
Consider a hypothesis set $H=\{h\}_i$, and let $h_\mc{S}^*\in \arg \min_{h\in H}\mc{L}_\mc{S}^c(h, l_\mc{S})$ and $h_\mc{T}^*\in \arg \min_{h\in H}\mc{L}_\mc{T}^c(h, l_\mc{T})$ be the classifiers that minimize the empirical risk on the source dataset $\mc{S}$ and the target dataset $\mc{T}$, respectively~\citep{zhang2020collaborative}.
The hypothesis error of the target domain classifier can be defined as $\epsilon_\mc{T}=\mc{L}_T^c(h,l_T)-\mc{L}_T^c(h_T^*,l_T)$. Since our difference distance loss function $disc_{\mc{L}}$ is symmetric and obeys the triangle inequality, according to domain adaptation theory \citep{mansour2009domain, ben2010theory} for any hypothesis $h\in H$, the following holds:
\begin{equation}
\begin{aligned}
    \mc{L}_\mc{T}^c(h,l_\mc{T})\leq \mc{L}_\mc{T}^c(h_\mc{T}^{*},l_\mc{T})+\mc{L}_S^c(h,h_\mc{S}^*)+ \\ 
    disc_{\mc{L}^d}(\mc{S},\mc{T})+\mc{L}_\mc{S}^c(h_\mc{S}^*,h_\mc{T}^*). 
\end{aligned}\tag{21}
\end{equation}
This inequality can be transformed into
\begin{equation}
    \epsilon_\mc{T} \leq \mc{L}_\mc{S}^c(h,h_\mc{S}^*)+disc_{\mc{L}^d}(\mc{S},\mc{T})+\mc{L}_\mc{S}^c(h_{S}^*,h_\mc{T}^*). \tag{22}
\end{equation}

We observe that the hypothesis error with respect to the target domain $\epsilon_{\mc{T}}$ is linked to the average loss of source classification  $\mc{L}_\mc{S}^c(h,h_S^*)$, the discrepancy distance $disc{\mc{L}}(\mc{S},\mc{T})$, and the average loss between the best intra-class hypotheses $\mc{L}_\mc{S}^c(h_\mc{S}^*,h_\mc{T}^*)$.

Since our method follows a two-stage training format, $\mc{L}_\mc{S}^c(h,h_S^*)$ remains constant after pre-training.
Furthermore, in order to achieve successful domain adaptation, it is reasonable for the optimal classifiers $h_\mc{S}^*$ and $h_\mc{T}^*$ in the source and target domains, respectively, to exhibit low inconsistency in semantic prediction, as measured by $\mc{L}_\mc{S}^c(h_\mc{S}^*,h_\mc{T}^*)$. 
With our proposed alignment loss $\mc{L}_{align}$, the model will force for samples in the target domain $\mc{T}$ to extract more features that express the style of the source domain $\mc{S}$, i.e., reduce $disc_{\mc{L}}(\mc{S}, \mc{T})$ to $disc_{\mc{L}}(\mc{S}, \mc{P})$ as the optimization proceeds. According to the conclusion of Proposition 1, this will reduce the target domain classifier error $\epsilon_\mathcal{T}$.$\hfill \square$



%
\balance

\bibliographystyle{spbasic}      
\bibliography{bibliography_file}

\begin{thebibliography}{80}
\providecommand{\natexlab}[1]{#1}
\providecommand{\url}[1]{{#1}}
\providecommand{\urlprefix}{URL }
\expandafter\ifx\csname urlstyle\endcsname\relax
  \providecommand{\doi}[1]{DOI~\discretionary{}{}{}#1}\else
  \providecommand{\doi}{DOI~\discretionary{}{}{}\begingroup \urlstyle{rm}\Url}\fi
\providecommand{\eprint}[2][]{\url{#2}}

\bibitem[{Ben-David et~al.(2010)Ben-David, Blitzer, Crammer, Kulesza, Pereira, and Vaughan}]{ben2010theory}
Ben-David S, Blitzer J, Crammer K, Kulesza A, Pereira F, Vaughan JW (2010) A theory of learning from different domains. Machine learning 79:151--175

\bibitem[{Bertinetto et~al.(2018)Bertinetto, Henriques, Torr, and Vedaldi}]{bertinetto2018meta}
Bertinetto L, Henriques JF, Torr P, Vedaldi A (2018) Meta-learning with differentiable closed-form solvers. In: International Conference on Learning Representations

\bibitem[{Chen et~al.(2019)Chen, Liu, Kira, Wang, and Huang}]{chen2019closer}
Chen WY, Liu YC, Kira Z, Wang YCF, Huang JB (2019) A closer look at few-shot classification. arXiv preprint arXiv:190404232

\bibitem[{Chen et~al.(2022)Chen, Rosenfeld, Sellke, Ma, and Risteski}]{chen2022iterative}
Chen Y, Rosenfeld E, Sellke M, Ma T, Risteski A (2022) Iterative feature matching: Toward provable domain generalization with logarithmic environments. Advances in Neural Information Processing Systems 35:1725--1736

\bibitem[{Cui et~al.(2020)Cui, Wang, Zhuo, Su, Huang, and Tian}]{cui2020gradually}
Cui S, Wang S, Zhuo J, Su C, Huang Q, Tian Q (2020) Gradually vanishing bridge for adversarial domain adaptation. In: Proceedings of the IEEE/CVF conference on computer vision and pattern recognition, pp 12455--12464

\bibitem[{Dai et~al.(2021)Dai, Liu, Sun, Tong, Zhang, and Duan}]{dai2021idm}
Dai Y, Liu J, Sun Y, Tong Z, Zhang C, Duan LY (2021) Idm: An intermediate domain module for domain adaptive person re-id. In: Proceedings of the IEEE/CVF International Conference on Computer Vision, pp 11864--11874

\bibitem[{Das et~al.(2022)Das, Yun, and Porikli}]{das2022confess}
Das D, Yun S, Porikli F (2022) Confess: A framework for single source cross-domain few-shot learning. In: International Conference on Learning Representations

\bibitem[{Das et~al.(2021)Das, Wang, and Moura}]{das2021importance}
Das R, Wang YX, Moura JM (2021) On the importance of distractors for few-shot classification. In: Proceedings of the IEEE/CVF International Conference on Computer Vision, pp 9030--9040

\bibitem[{Deng et~al.(2009)Deng, Dong, Socher, Li, Li, and Fei-Fei}]{deng2009imagenet}
Deng J, Dong W, Socher R, Li LJ, Li K, Fei-Fei L (2009) Imagenet: A large-scale hierarchical image database. In: 2009 IEEE conference on computer vision and pattern recognition, Ieee, pp 248--255

\bibitem[{Doersch et~al.(2020)Doersch, Gupta, and Zisserman}]{doersch2020crosstransformers}
Doersch C, Gupta A, Zisserman A (2020) Crosstransformers: spatially-aware few-shot transfer. Advances in Neural Information Processing Systems 33:21981--21993

\bibitem[{Fei-Fei et~al.(2006)Fei-Fei, Fergus, and Perona}]{fei2006one}
Fei-Fei L, Fergus R, Perona P (2006) One-shot learning of object categories. IEEE transactions on pattern analysis and machine intelligence 28(4):594--611

\bibitem[{Finn et~al.(2017)Finn, Abbeel, and Levine}]{finn2017model}
Finn C, Abbeel P, Levine S (2017) Model-agnostic meta-learning for fast adaptation of deep networks. In: International conference on machine learning, PMLR, pp 1126--1135

\bibitem[{Ganin et~al.(2016)Ganin, Ustinova, Ajakan, Germain, Larochelle, Laviolette, Marchand, and Lempitsky}]{ganin2016domain}
Ganin Y, Ustinova E, Ajakan H, Germain P, Larochelle H, Laviolette F, Marchand M, Lempitsky V (2016) Domain-adversarial training of neural networks. The journal of machine learning research 17(1):2096--2030

\bibitem[{Garcia and Bruna(2018)}]{garcia2018few}
Garcia V, Bruna J (2018) Few-shot learning with graph neural networks. In: 6th International Conference on Learning Representations, ICLR 2018

\bibitem[{Gatys et~al.(2015)Gatys, Ecker, and Bethge}]{gatys2015neural}
Gatys LA, Ecker AS, Bethge M (2015) A neural algorithm of artistic style. arXiv preprint arXiv:150806576

\bibitem[{Gidaris et~al.(2019)Gidaris, Bursuc, Komodakis, P{\'e}rez, and Cord}]{gidaris2019boosting}
Gidaris S, Bursuc A, Komodakis N, P{\'e}rez P, Cord M (2019) Boosting few-shot visual learning with self-supervision. In: Proceedings of the IEEE/CVF international conference on computer vision, pp 8059--8068

\bibitem[{Gong et~al.(2012)Gong, Shi, Sha, and Grauman}]{gong2012geodesic}
Gong B, Shi Y, Sha F, Grauman K (2012) Geodesic flow kernel for unsupervised domain adaptation. In: 2012 IEEE conference on computer vision and pattern recognition, IEEE, pp 2066--2073

\bibitem[{Gopalan et~al.(2013)Gopalan, Li, and Chellappa}]{gopalan2013unsupervised}
Gopalan R, Li R, Chellappa R (2013) Unsupervised adaptation across domain shifts by generating intermediate data representations. IEEE transactions on pattern analysis and machine intelligence 36(11):2288--2302

\bibitem[{Guo et~al.(2022)Guo, Tian, Zhao, Zhou, and Zha}]{guo2022adaptive}
Guo D, Tian L, Zhao H, Zhou M, Zha H (2022) Adaptive distribution calibration for few-shot learning with hierarchical optimal transport. Advances in neural information processing systems 35:6996--7010

\bibitem[{Guo et~al.(2020)Guo, Codella, Karlinsky, Codella, Smith, Saenko, Rosing, and Feris}]{guo2020broader}
Guo Y, Codella NC, Karlinsky L, Codella JV, Smith JR, Saenko K, Rosing T, Feris R (2020) A broader study of cross-domain few-shot learning. In: European conference on computer vision, Springer, pp 124--141

\bibitem[{He et~al.(2016)He, Zhang, Ren, and Sun}]{he2016deep}
He K, Zhang X, Ren S, Sun J (2016) Deep residual learning for image recognition. In: Proceedings of the IEEE conference on computer vision and pattern recognition, pp 770--778

\bibitem[{Helber et~al.(2019)Helber, Bischke, Dengel, and Borth}]{helber2019eurosat}
Helber P, Bischke B, Dengel A, Borth D (2019) Eurosat: A novel dataset and deep learning benchmark for land use and land cover classification. IEEE Journal of Selected Topics in Applied Earth Observations and Remote Sensing 12(7):2217--2226

\bibitem[{Hoerl and Kennard(1970)}]{hoerl1970ridge}
Hoerl AE, Kennard RW (1970) Ridge regression: Biased estimation for nonorthogonal problems. Technometrics 12(1):55--67

\bibitem[{Hu and Ma(2022)}]{hu2022adversarial}
Hu Y, Ma AJ (2022) Adversarial feature augmentation for cross-domain few-shot classification. In: Computer Vision--ECCV 2022: 17th European Conference, Tel Aviv, Israel, October 23--27, 2022, Proceedings, Part XX, Springer, pp 20--37

\bibitem[{Huang and Choi(2023)}]{huang2023sapenet}
Huang X, Choi SH (2023) Sapenet: Self-attention based prototype enhancement network for few-shot learning. Pattern Recognition 135:109170

\bibitem[{Ioffe and Szegedy(2015)}]{ioffe2015batch}
Ioffe S, Szegedy C (2015) Batch normalization: Accelerating deep network training by reducing internal covariate shift. In: International conference on machine learning, pmlr, pp 448--456

\bibitem[{Kang et~al.(2018)Kang, Zheng, Yan, and Yang}]{kang2018deep}
Kang G, Zheng L, Yan Y, Yang Y (2018) Deep adversarial attention alignment for unsupervised domain adaptation: the benefit of target expectation maximization. In: Proceedings of the European conference on computer vision (ECCV), pp 401--416

\bibitem[{Kemker et~al.(2018)Kemker, McClure, Abitino, Hayes, and Kanan}]{kemker2018measuring}
Kemker R, McClure M, Abitino A, Hayes TL, Kanan C (2018) Measuring catastrophic forgetting in neural networks. In: Thirty-Second AAAI Conference on Artificial Intelligence

\bibitem[{Krause et~al.(2013)Krause, Stark, Deng, and Fei-Fei}]{krause20133d}
Krause J, Stark M, Deng J, Fei-Fei L (2013) 3d object representations for fine-grained categorization. In: Proceedings of the IEEE international conference on computer vision workshops, pp 554--561

\bibitem[{Lake et~al.(2015)Lake, Salakhutdinov, and Tenenbaum}]{lake2015human}
Lake BM, Salakhutdinov R, Tenenbaum JB (2015) Human-level concept learning through probabilistic program induction. Science 350(6266):1332--1338

\bibitem[{Li et~al.(2022{\natexlab{a}})Li, Gong, Wang, and Fu}]{li2022ranking}
Li P, Gong S, Wang C, Fu Y (2022{\natexlab{a}}) Ranking distance calibration for cross-domain few-shot learning. In: Proceedings of the IEEE/CVF Conference on Computer Vision and Pattern Recognition, pp 9099--9108

\bibitem[{Li et~al.(2023)Li, Liu, Jiao, Li, Li, Liu, and Huang}]{li2023knowledge}
Li P, Liu F, Jiao L, Li S, Li L, Liu X, Huang X (2023) Knowledge transduction for cross-domain few-shot learning. Pattern Recognition 141:109652

\bibitem[{Li et~al.(2022{\natexlab{b}})Li, Liu, and Bilen}]{li2022cross}
Li WH, Liu X, Bilen H (2022{\natexlab{b}}) Cross-domain few-shot learning with task-specific adapters. In: Proceedings of the IEEE/CVF conference on computer vision and pattern recognition, pp 7161--7170

\bibitem[{Li et~al.(2016)Li, Wang, Shi, Liu, and Hou}]{li2016revisiting}
Li Y, Wang N, Shi J, Liu J, Hou X (2016) Revisiting batch normalization for practical domain adaptation. arXiv preprint arXiv:160304779

\bibitem[{Li et~al.(2017)Li, Zhou, Chen, and Li}]{li2017meta}
Li Z, Zhou F, Chen F, Li H (2017) Meta-sgd: Learning to learn quickly for few-shot learning. arXiv preprint arXiv:170709835

\bibitem[{Liang et~al.(2021)Liang, Zhang, Dai, and Lu}]{liang2021boosting}
Liang H, Zhang Q, Dai P, Lu J (2021) Boosting the generalization capability in cross-domain few-shot learning via noise-enhanced supervised autoencoder. In: Proceedings of the IEEE/CVF International Conference on Computer Vision, pp 9424--9434

\bibitem[{Liu et~al.(2020)Liu, Zhao, Li, Jiang, Guo, and Ye}]{liu2020feature}
Liu B, Zhao Z, Li Z, Jiang J, Guo Y, Ye J (2020) Feature transformation ensemble model with batch spectral regularization for cross-domain few-shot classification. arXiv preprint arXiv:200508463

\bibitem[{Luo et~al.(2021)Luo, Wei, Wen, Yang, Xie, Xu, and Tian}]{luo2021rectifying}
Luo X, Wei L, Wen L, Yang J, Xie L, Xu Z, Tian Q (2021) Rectifying the shortcut learning of background for few-shot learning. Advances in Neural Information Processing Systems 34:13073--13085

\bibitem[{Mairal et~al.(2010)Mairal, Bach, Ponce, and Sapiro}]{mairal2010online}
Mairal J, Bach F, Ponce J, Sapiro G (2010) Online learning for matrix factorization and sparse coding. Journal of Machine Learning Research 11(1)

\bibitem[{Mansour et~al.(2009)Mansour, Mohri, and Rostamizadeh}]{mansour2009domain}
Mansour Y, Mohri M, Rostamizadeh A (2009) Domain adaptation: Learning bounds and algorithms. arXiv preprint arXiv:09023430

\bibitem[{Maria~Carlucci et~al.(2017)Maria~Carlucci, Porzi, Caputo, Ricci, and Rota~Bulo}]{maria2017autodial}
Maria~Carlucci F, Porzi L, Caputo B, Ricci E, Rota~Bulo S (2017) Autodial: Automatic domain alignment layers. In: Proceedings of the IEEE international conference on computer vision, pp 5067--5075

\bibitem[{McCloskey and Cohen(1989)}]{mccloskey1989catastrophic}
McCloskey M, Cohen NJ (1989) Catastrophic interference in connectionist networks: The sequential learning problem. In: Psychology of learning and motivation, vol~24, Elsevier, pp 109--165

\bibitem[{Miller et~al.(2000)Miller, Matsakis, and Viola}]{miller2000learning}
Miller EG, Matsakis NE, Viola PA (2000) Learning from one example through shared densities on transforms. In: Proceedings IEEE Conference on Computer Vision and Pattern Recognition. CVPR 2000 (Cat. No. PR00662), IEEE, vol~1, pp 464--471

\bibitem[{Nichol et~al.(2018)Nichol, Achiam, and Schulman}]{nichol2018first}
Nichol A, Achiam J, Schulman J (2018) On first-order meta-learning algorithms. arXiv preprint arXiv:180302999

\bibitem[{Oreshkin et~al.(2018)Oreshkin, Rodr{\'\i}guez~L{\'o}pez, and Lacoste}]{oreshkin2018tadam}
Oreshkin B, Rodr{\'\i}guez~L{\'o}pez P, Lacoste A (2018) Tadam: Task dependent adaptive metric for improved few-shot learning. Advances in neural information processing systems 31

\bibitem[{Pan et~al.(2010)Pan, Tsang, Kwok, and Yang}]{pan2010domain}
Pan SJ, Tsang IW, Kwok JT, Yang Q (2010) Domain adaptation via transfer component analysis. IEEE transactions on neural networks 22(2):199--210

\bibitem[{Paszke et~al.(2019)Paszke, Gross, Massa, Lerer, Bradbury, Chanan, Killeen, Lin, Gimelshein, Antiga et~al.}]{paszke2019pytorch}
Paszke A, Gross S, Massa F, Lerer A, Bradbury J, Chanan G, Killeen T, Lin Z, Gimelshein N, Antiga L, et~al. (2019) Pytorch: An imperative style, high-performance deep learning library. Advances in neural information processing systems 32

\bibitem[{Phoo and Hariharan(2020)}]{phoo2020self}
Phoo CP, Hariharan B (2020) Self-training for few-shot transfer across extreme task differences. arXiv preprint arXiv:201007734

\bibitem[{Rajpurkar et~al.(2017)Rajpurkar, Irvin, Zhu, Yang, Mehta, Duan, Ding, Bagul, Langlotz, Shpanskaya et~al.}]{rajpurkar2017chexnet}
Rajpurkar P, Irvin J, Zhu K, Yang B, Mehta H, Duan T, Ding D, Bagul A, Langlotz C, Shpanskaya K, et~al. (2017) Chexnet: Radiologist-level pneumonia detection on chest x-rays with deep learning. arXiv preprint arXiv:171105225

\bibitem[{Rizve et~al.(2021)Rizve, Khan, Khan, and Shah}]{rizve2021exploring}
Rizve MN, Khan S, Khan FS, Shah M (2021) Exploring complementary strengths of invariant and equivariant representations for few-shot learning. In: Proceedings of the IEEE/CVF Conference on Computer Vision and Pattern Recognition, pp 10836--10846

\bibitem[{Robey et~al.(2021)Robey, Pappas, and Hassani}]{robey2021model}
Robey A, Pappas GJ, Hassani H (2021) Model-based domain generalization. Advances in Neural Information Processing Systems 34:20210--20229

\bibitem[{Rusu et~al.(2018)Rusu, Rao, Sygnowski, Vinyals, Pascanu, Osindero, and Hadsell}]{rusu2018meta}
Rusu AA, Rao D, Sygnowski J, Vinyals O, Pascanu R, Osindero S, Hadsell R (2018) Meta-learning with latent embedding optimization. arXiv preprint arXiv:180705960

\bibitem[{Shirekar et~al.(2023)Shirekar, Singh, and Jamali-Rad}]{shirekar2023self}
Shirekar OK, Singh A, Jamali-Rad H (2023) Self-attention message passing for contrastive few-shot learning. In: Proceedings of the IEEE/CVF Winter Conference on Applications of Computer Vision, pp 5426--5436

\bibitem[{Simonyan and Zisserman(2014)}]{simonyan2014very}
Simonyan K, Zisserman A (2014) Very deep convolutional networks for large-scale image recognition. arXiv preprint arXiv:14091556

\bibitem[{Snell et~al.(2017)Snell, Swersky, and Zemel}]{snell2017prototypical}
Snell J, Swersky K, Zemel R (2017) Prototypical networks for few-shot learning. Advances in neural information processing systems 30

\bibitem[{Sun et~al.(2021)Sun, Lapuschkin, Samek, Zhao, Cheung, and Binder}]{sun2021explanation}
Sun J, Lapuschkin S, Samek W, Zhao Y, Cheung NM, Binder A (2021) Explanation-guided training for cross-domain few-shot classification. In: 2020 25th International Conference on Pattern Recognition (ICPR), IEEE, pp 7609--7616

\bibitem[{Sung et~al.(2018)Sung, Yang, Zhang, Xiang, Torr, and Hospedales}]{sung2018learning}
Sung F, Yang Y, Zhang L, Xiang T, Torr PH, Hospedales TM (2018) Learning to compare: Relation network for few-shot learning. In: Proceedings of the IEEE conference on computer vision and pattern recognition, pp 1199--1208

\bibitem[{Triantafillou et~al.(2019)Triantafillou, Zhu, Dumoulin, Lamblin, Evci, Xu, Goroshin, Gelada, Swersky, Manzagol et~al.}]{triantafillou2019meta}
Triantafillou E, Zhu T, Dumoulin V, Lamblin P, Evci U, Xu K, Goroshin R, Gelada C, Swersky K, Manzagol PA, et~al. (2019) Meta-dataset: A dataset of datasets for learning to learn from few examples. In: International Conference on Learning Representations

\bibitem[{Tseng et~al.(2020)Tseng, Lee, Huang, and Yang}]{tseng2020cross}
Tseng HY, Lee HY, Huang JB, Yang MH (2020) Cross-domain few-shot classification via learned feature-wise transformation. arXiv preprint arXiv:200108735

\bibitem[{Tzeng et~al.(2014)Tzeng, Hoffman, Zhang, Saenko, and Darrell}]{tzeng2014deep}
Tzeng E, Hoffman J, Zhang N, Saenko K, Darrell T (2014) Deep domain confusion: Maximizing for domain invariance. arXiv preprint arXiv:14123474

\bibitem[{Tzeng et~al.(2017)Tzeng, Hoffman, Saenko, and Darrell}]{tzeng2017adversarial}
Tzeng E, Hoffman J, Saenko K, Darrell T (2017) Adversarial discriminative domain adaptation. In: Proceedings of the IEEE conference on computer vision and pattern recognition, pp 7167--7176

\bibitem[{Van~Horn et~al.(2018)Van~Horn, Mac~Aodha, Song, Cui, Sun, Shepard, Adam, Perona, and Belongie}]{van2018inaturalist}
Van~Horn G, Mac~Aodha O, Song Y, Cui Y, Sun C, Shepard A, Adam H, Perona P, Belongie S (2018) The inaturalist species classification and detection dataset. In: Proceedings of the IEEE conference on computer vision and pattern recognition, pp 8769--8778

\bibitem[{Vinyals et~al.(2016)Vinyals, Blundell, Lillicrap, Wierstra et~al.}]{vinyals2016matching}
Vinyals O, Blundell C, Lillicrap T, Wierstra D, et~al. (2016) Matching networks for one shot learning. Advances in neural information processing systems 29

\bibitem[{Vuorio et~al.(2019)Vuorio, Sun, Hu, and Lim}]{vuorio2019multimodal}
Vuorio R, Sun SH, Hu H, Lim JJ (2019) Multimodal model-agnostic meta-learning via task-aware modulation. Advances in Neural Information Processing Systems 32

\bibitem[{Wah et~al.(2011)Wah, Branson, Welinder, Perona, and Belongie}]{wah2011caltech}
Wah C, Branson S, Welinder P, Perona P, Belongie S (2011) The caltech-ucsd birds-200-2011 dataset. Tech. rep., California Institute of Technology

\bibitem[{Wang and Deng(2021{\natexlab{a}})}]{wang2021cross}
Wang H, Deng ZH (2021{\natexlab{a}}) Cross-domain few-shot classification via adversarial task augmentation. arXiv preprint arXiv:210414385

\bibitem[{Wang and Deng(2021{\natexlab{b}})}]{ijcai2021-149}
Wang H, Deng ZH (2021{\natexlab{b}}) Cross-domain few-shot classification via adversarial task augmentation. In: Zhou ZH (ed) Proceedings of the Thirtieth International Joint Conference on Artificial Intelligence, {IJCAI-21}, International Joint Conferences on Artificial Intelligence Organization, pp 1075--1081, main Track

\bibitem[{Wang et~al.(2017)Wang, Peng, Lu, Lu, Bagheri, and Summers}]{wang2017chestx}
Wang X, Peng Y, Lu L, Lu Z, Bagheri M, Summers RM (2017) Chestx-ray8: Hospital-scale chest x-ray database and benchmarks on weakly-supervised classification and localization of common thorax diseases. In: Proceedings of the IEEE conference on computer vision and pattern recognition, pp 2097--2106

\bibitem[{Wang et~al.(2019)Wang, Jin, Long, Wang, and Jordan}]{wang2019transferable}
Wang X, Jin Y, Long M, Wang J, Jordan MI (2019) Transferable normalization: Towards improving transferability of deep neural networks. Advances in neural information processing systems 32

\bibitem[{Wei et~al.(2022)Wei, Xu, Zhang, Peng, and Zhou}]{wei2022embarrassingly}
Wei XS, Xu HY, Zhang F, Peng Y, Zhou W (2022) An embarrassingly simple approach to semi-supervised few-shot learning. Advances in Neural Information Processing Systems 35:14489--14500

\bibitem[{Wertheimer et~al.(2021)Wertheimer, Tang, and Hariharan}]{wertheimer2021few}
Wertheimer D, Tang L, Hariharan B (2021) Few-shot classification with feature map reconstruction networks. In: Proceedings of the IEEE/CVF Conference on Computer Vision and Pattern Recognition, pp 8012--8021

\bibitem[{Xu et~al.(2022{\natexlab{a}})Xu, Luo, Pan, Li, Pei, and Xu}]{xu2022alleviating}
Xu J, Luo X, Pan X, Li Y, Pei W, Xu Z (2022{\natexlab{a}}) Alleviating the sample selection bias in few-shot learning by removing projection to the centroid. Advances in Neural Information Processing Systems 35:21073--21086

\bibitem[{Xu et~al.(2022{\natexlab{b}})Xu, Wang, Wang, Qin, Zhang, and Fu}]{xu2021memrein}
Xu Y, Wang L, Wang Y, Qin C, Zhang Y, Fu Y (2022{\natexlab{b}}) Memrein: Rein the domain shift for cross-domain few-shot learning. In: Proceedings of the Thirty-First International Joint Conference on Artificial Intelligence, {IJCAI} 2022, Vienna, Austria, 23-29 July 2022, pp 3636--3642

\bibitem[{Ye et~al.(2020)Ye, Hu, Zhan, and Sha}]{ye2020few}
Ye HJ, Hu H, Zhan DC, Sha F (2020) Few-shot learning via embedding adaptation with set-to-set functions. In: Proceedings of the IEEE/CVF Conference on Computer Vision and Pattern Recognition, pp 8808--8817

\bibitem[{Zhang et~al.(2020{\natexlab{a}})Zhang, Cai, Lin, and Shen}]{zhang2020deepemd}
Zhang C, Cai Y, Lin G, Shen C (2020{\natexlab{a}}) Deepemd: Few-shot image classification with differentiable earth mover's distance and structured classifiers. In: Proceedings of the IEEE/CVF conference on computer vision and pattern recognition, pp 12203--12213

\bibitem[{Zhang et~al.(2020{\natexlab{b}})Zhang, Qi, Shi, and Gao}]{zhang2020generalizable}
Zhang J, Qi L, Shi Y, Gao Y (2020{\natexlab{b}}) Generalizable semantic segmentation via model-agnostic learning and target-specific normalization. arXiv preprint arXiv:200312296 2(3):6

\bibitem[{Zhang et~al.(2019)Zhang, Liu, Long, and Jordan}]{zhang2019bridging}
Zhang Y, Liu T, Long M, Jordan M (2019) Bridging theory and algorithm for domain adaptation. In: International conference on machine learning, PMLR, pp 7404--7413

\bibitem[{Zhang et~al.(2020{\natexlab{c}})Zhang, Wei, Wu, Zhao, Niu, Huang, and Tan}]{zhang2020collaborative}
Zhang Y, Wei Y, Wu Q, Zhao P, Niu S, Huang J, Tan M (2020{\natexlab{c}}) Collaborative unsupervised domain adaptation for medical image diagnosis. IEEE Transactions on Image Processing 29:7834--7844

\bibitem[{Zhou et~al.(2017)Zhou, Lapedriza, Khosla, Oliva, and Torralba}]{zhou2017places}
Zhou B, Lapedriza A, Khosla A, Oliva A, Torralba A (2017) Places: A 10 million image database for scene recognition. IEEE transactions on pattern analysis and machine intelligence 40(6):1452--1464

\bibitem[{Zhou et~al.(2023)Zhou, Wang, Zhang, Wei, and Zhang}]{zhou2023revisiting}
Zhou F, Wang P, Zhang L, Wei W, Zhang Y (2023) Revisiting prototypical network for cross-domain few-shot learning. In: Proceedings of the IEEE/CVF Conference on Computer Vision and Pattern Recognition, pp 20061--20070

\end{thebibliography}

\section*{Data Availability Statement}

The datasets generated during and/or analyzed during the current research are publicly available in the following references,~\ie, ImageNet~\citep{deng2009imagenet}\url{https://www.image-net.org/}, Stanford Cars~\citep{krause20133d}~\url{https://ai.stanford.edu/~jkrause/cars/car_dataset.html}, CUB-200-2011~\citep{wah2011caltech}~\url{https://www.vision.caltech.edu/datasets/cub_200_2011/}, Plantae~\citep{van2018inaturalist}~\url{https://github.com/visipedia/inat_comp/tree/master/2017} and Places datasets~\citep{zhou2017places}~\url{http://places.csail.mit.edu/}. Diverse domain benchmarking for CropDisease, EuroSAT, ISIC, and ChestX are included in BSCD-FSL benchmark~\citep{guo2020broader}~\url{https://github.com/IBM/cdfsl-benchmark}.
The source codes and models corresponding to this study are publicly available.

\end{document}